%% file: main.tex
\documentclass[10pt]{article}
\usepackage[utf8]{inputenc}
\usepackage[T1]{fontenc}
\usepackage{amsmath}
\usepackage{amssymb}
\usepackage{pifont}
\usepackage{graphicx}
\usepackage{url}
\usepackage{subcaption}
\usepackage{float}
\usepackage{xcolor}
\usepackage{colortbl}
\usepackage{array}
\usepackage{tabularx}
\usepackage{booktabs}
\usepackage{multirow}
\usepackage{makecell}
\usepackage{geometry}
\usepackage{longtable}
\usepackage{pdflscape}
\usepackage{listings}
\usepackage{tcolorbox}
\tcbuselibrary{skins}
\usepackage{tikz}
\usetikzlibrary{arrows.meta,positioning,calc,shapes.geometric}
\usepackage{pgfplots}
\pgfplotsset{compat=1.18}
\usepackage{fancyhdr}
\usepackage[colorlinks=true,linkcolor=black,citecolor=black,urlcolor=black,linktoc=all]{hyperref}
\usepackage{bookmark}

\newcommand{\paperauthorfont}{\small}

\definecolor{wenxinblue}{HTML}{3d58db}
\definecolor{wenxinbg}{HTML}{f0f6ff}
\definecolor{baidublue}{HTML}{0a74f9}
\definecolor{baidured}{HTML}{dd3944}
\hypersetup{colorlinks=true, citecolor=wenxinblue, linkcolor=wenxinblue, urlcolor=wenxinblue}

\newcommand{\paperdate}{\today}

\definecolor{limitred}{RGB}{220,53,69}
\definecolor{taskblue}{RGB}{0,123,255}
\definecolor{spatorange}{RGB}{253,126,20}
\definecolor{avpurple}{RGB}{111,66,193}
\definecolor{metricgreen}{RGB}{40,167,69}
\definecolor{moonveil}{RGB}{248,249,250}

\newcommand{\xmark}{\textcolor{red}{\ding{55}}} 
\newsavebox{\minervabox}

\makeatletter
\newcommand{\mymaketitle}{%
\noindent
{\color{black}\hrule height 0.7pt}\par
\vspace{1.35em}%
{\Large\bfseries \centering \@title \par}
\vspace{1.35em}%
{\color{black}\hrule height 0.7pt}\par
\vspace{0.85em}%
{\paperauthorfont \centering \@author \par}
\vspace{1.0em}
  \begin{tcolorbox}[
    enhanced, frame hidden,
    colback=wenxinbg,
    arc=7pt,
    boxrule=0pt,
    left=0.7cm,
    right=0.7cm,
    top=0.5cm,
    bottom=0.4cm,
    before skip=0pt,
    after skip=0.3cm,
  ]
    {\normalsize\bfseries Abstract\par}
    \vspace{0.4em}
    {\small \@abstract\par}
    \vspace{0.45em}
    {\small \textbf{Date: }\paperdate\par}
    \vspace{0.25em}
    {\small \textbf{Project page: }\url{https://go-agent-x.github.io/video_agent_harness/}\par}
  \end{tcolorbox}
  \thispagestyle{plain}%
}
\renewcommand{\maketitle}{\par\mymaketitle}

\fancypagestyle{plain}{%
  \fancyhf{}%
  \cfoot{\thepage}%
}
\makeatother

\begin{document}

\title{Online Video Agent Harness for Long Video Understanding}
\author{Sen Yang\textsuperscript{*}, Boqiang Duan\textsuperscript{*}, Jing Yang\textsuperscript{\dag}, Weihao Bo\textsuperscript{\dag}, Jie Liu\textsuperscript{\dag}, Boyuan Tong\textsuperscript{\dag}, Ze Feng\textsuperscript{\dag}, Wenkang Zhang\textsuperscript{\dag},\\ Jingdong Wang, Hua Wu \\[0.8em]
Baidu Inc.}
\date{}
\makeatletter
\def\@abstract{%
Long video understanding often behaves like a visual needle-in-a-haystack problem: query-relevant evidence is sparsely distributed across long temporal spans, while packing dense frames into a single VLM context incurs \textit{context rot} and high cost.
Existing video agents often rely on query-agnostic offline preprocessing or ad hoc tool sets, which can miss query-specific details and waste computation.
In this work, we present VideoXAgent, a purely online video-agent harness for long video understanding that starts from the given video file and user query, plans and decomposes the task, invokes specialized expert tools on demand, and aggregates multimodal evidence to produce a final answer while resolving conflicts among observations.
To support this on-demand invocation, we design a suite of heterogeneous expert tools guided by a data-driven taxonomy of atomic capabilities, spanning scripts, VLMs, and domain models (e.g., detection, OCR, ASR, face recognition).
The harness further enforces objective evidence prompting and budget-aware control to curb hallucination and non-termination.
Across Video-MME-Long, LongVideoBench-Long, LVBench, and MINERVA, VideoXAgent is competitive with frontier LMMs and video agents under a smaller context footprint---about 50k tokens of agent context per sample, even on hour-long videos.
In particular, on complex video-reasoning benchmarks such as MINERVA, it matches this level while using only about 15\% of the context of a 1,024-frame dense-packing baseline.
Notably, the harness remains effective with a visually weak or even text-only orchestrator, suggesting that strong long-video understanding can emerge from progressive agentic evidence seeking rather than from packing the full video into a single context.
}
\makeatother
\maketitle
\begingroup
\renewcommand{\thefootnote}{\fnsymbol{footnote}}
\footnotetext[1]{Equal contributions. Contact Email: yangsenius@gmail.com, firestonelib@gmail}
\footnotetext[2]{The work was done when Jing Yang, Weihao Bo, Jie Liu, Boyuan Tong, Ze Feng, and Wenkang Zhang were interns at Baidu.}
\endgroup
\begin{figure}[!h]
  \centering
  \includegraphics[width=\textwidth,height=0.24\textheight,keepaspectratio]{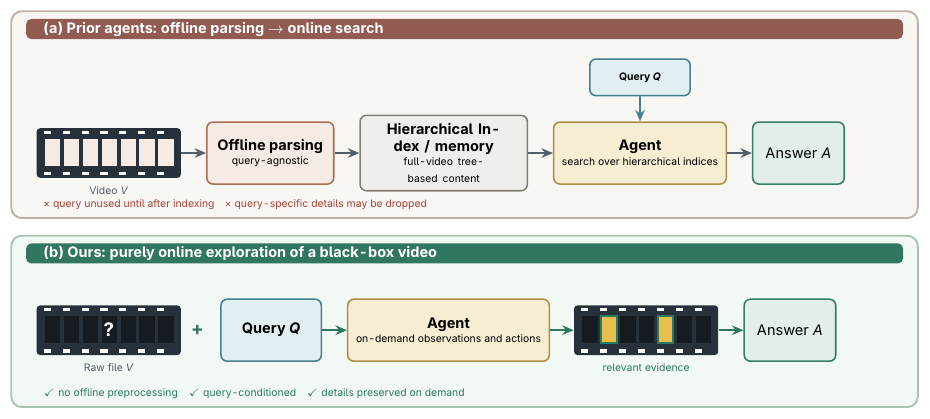}
  \caption{Two video-agent paradigms. (a)~Prior agents first run a query-agnostic preprocessing stage: they split the full video into fixed-length segments, uniformly sample frames, and parse it offline into an intermediate database or hierarchical memory. Query-specific reasoning and retrieval begin only after this index exists, so evidence not preserved by the fixed sampling can be missed, and a full-video cost is incurred before any query. (b)~Our purely online agent starts from the raw video file together with the user query, and acquires task-relevant evidence on demand by invoking tools on query-conditioned spans.}
  \label{fig:offline_vs_online}
\end{figure}
\clearpage
\pagestyle{plain}

\section{Introduction}

Long video understanding is a key multimodal application and a practical test of the long-context capabilities of large models.
A long-form video—such as a short film, animation, vlog, movie, or any other kind of video recording—may run from tens of minutes to several hours. Even with sparse sampling at 1 FPS, such videos can contain thousands of frames. Under the standard processing paradigm for large multimodal models (LMMs), this can easily lead to contexts with millions of tokens, far beyond the effective context length of most models and thus prone to context degradation~\cite{hong2025context, wu2025visual}.
Existing model-based solutions typically rely on direct model understanding: dynamically sampled or low-resolution frames are packed into the context window to provide broad temporal coverage, as in methods such as Qwen2.5/3-VL~\cite{qwenvl:bai2025qwen25vl,qwenvl:bai2025qwen3vl} and Kimi-K2.5~\cite{kimi:team2026k25}.
However, when the distribution of relevant information in the video cannot be assumed in advance, uniform downsampling has two major drawbacks: (1) it may miss critical frames that contain query-relevant evidence; (2) it also introduces many irrelevant frames from unrelated temporal segments, which can further interfere with visual reasoning.
Some recent methods, such as VideoThinker~\cite{videothinker:li2026agentic}, LongVT~\cite{longvt:yang2026thinking}, and Doubao-Seed-1.8/2.0~\cite{seed:team2026seed18,seed:team2026seed20}, attempt to first reason about potentially relevant temporal intervals and then perform dense sampling within those segments to inspect details and identify evidence.
Nevertheless, these methods still mostly pack frames into one model context.
They make limited use of non-visual signals such as speech, on-screen text, and video structure, and do not treat long video understanding as a multi-step process of seeking evidence and jointly reasoning across modalities.

In this context, it is natural to use the agentic paradigm to solve long video understanding, as the agent can adaptively interpret both the video content and the user query, decompose the problem into intermediate subproblems, and selectively invoke specialized tools to acquire task-relevant evidence.
However, many existing video-agent systems, such as DVD~\cite{dvd:zhang2026deep} and HAVEN~\cite{haven:yin2026hierarchical}, still rely on a query-agnostic preprocessing stage before agentic reasoning begins.
In such a stage, they usually split the full video into fixed-length segments, uniformly sample frames, combine audio content, and parse the video offline into an intermediate database or hierarchical memory.
Only after this precomputation step do they perform query-specific agentic reasoning and retrieval. Nevertheless, this two-stage pipeline still inherits the limitations of fixed sampling, can overlook query-specific evidence that was not preserved during preprocessing, and incurs substantial computational overhead, particularly for long videos.

In this work, we propose a purely online agentic paradigm for long video understanding, together with a principled agent harness that follows contemporary agent-design principles: the agent adaptively reasons over the video content and the user query, decomposes the task into multiple continuous steps, and selectively invokes external tools to acquire task-relevant evidence.
Within this agentic paradigm, the orchestrated models play a fundamentally different role from VLMs in conventional video-understanding pipelines.
Instead of loading the entire video context at once, they act as adaptive reasoners, treating the video as a black box and progressively disclosing evidence via intermediate reasoning and on-demand tool invocation.
This interactive and query-conditioned process is particularly suitable for long videos, where the information required to answer a question is often sparsely distributed across time and may involve heterogeneous modalities, including visual events, speech, text, faces, objects, and temporal relations.
Figure~\ref{fig:offline_vs_online} contrasts this offline-then-online pipeline with a purely online alternative that never materializes a query-agnostic index.

As for the agent orchestration process, we also consider the problem from another perspective: the capabilities that video understanding requires, and the corresponding tool design.
Current video-agent designs, however, often choose those tools in an ad hoc manner, based mainly on human intuition.
Although such designs can be effective for selected scenarios, they lack a systematic account of what capabilities are intrinsically required by complex and long-tail video understanding scenes.
As a result, existing agents may include redundant tools, miss important atomic capabilities, or provide tools whose interfaces are too ambiguous for reliable orchestration. This suggests that building a strong and online long video agent requires not only adding more tools, but also designing a principled tool space and a specialized agent harness that reflect the underlying capability structure of video understanding.

The challenge is that a query rarely requires a single capability. Depending on the video and the question, answering it may require temporal or spatial localization, audio understanding, and logical inference. These underlying capabilities are difficult to annotate manually or infer reliably from video--question pairs. Existing benchmarks mainly label question formats, genres, or coarse task types, and therefore provide little guidance for decomposing a question into atomic capabilities.
To address this gap, rather than subjectively defining the required capabilities, we empirically discover a comprehensive taxonomy of atomic capabilities directly from the reasoning processes of human experts.
We then use this taxonomy as a blueprint for expert-tool design, mapping each capability to specialized tools spanning scripts, VLMs, and domain models.
The online agent decomposes the query over this tool space and acquires evidence only as needed.

\begin{figure}[t]
\centering
\begin{minipage}{0.48\textwidth}
    \centering
    \includegraphics[width=\linewidth]{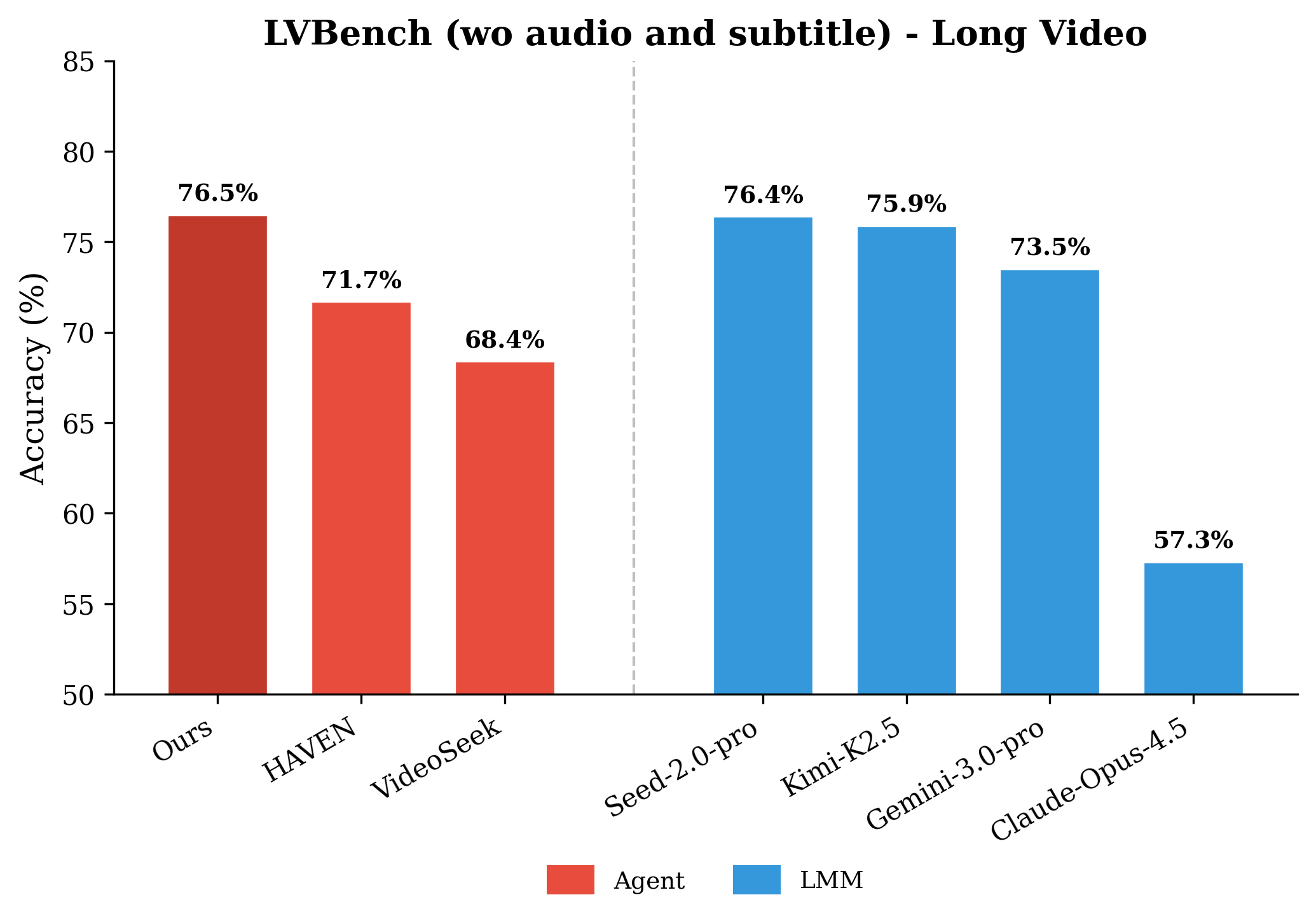}
\end{minipage}
\hfill
\begin{minipage}{0.48\textwidth}
    \centering
    \includegraphics[width=\linewidth]{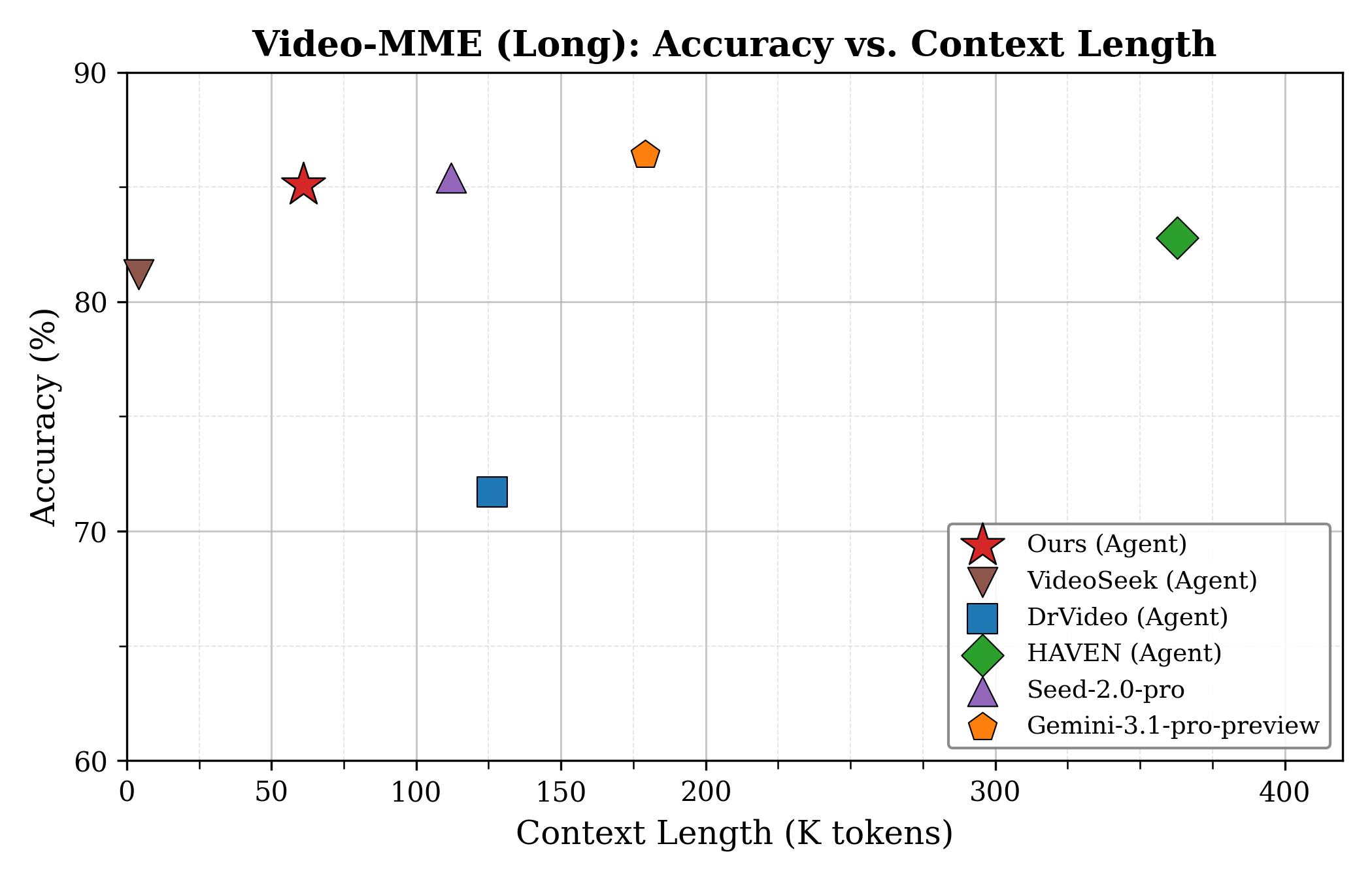}
\end{minipage}
\caption{Accuracy and efficiency comparison on long video understanding benchmarks. Left: LVBench accuracy across representative methods; the HAVEN value is the visual-only ablation without subtitles or audio. Right: Video-MME-Long accuracy against input-context cost. For our method, context length includes the total agent context and the visual tokens of frames processed by the VLM tools for a single video; for the remaining models, it is estimated from the reported number of processed frames at 256 tokens per frame.}
\label{fig:intro_accuracy_efficiency}
\end{figure}

Figure~\ref{fig:intro_accuracy_efficiency} shows the performance and efficiency of this design.
Under the online, taxonomy-guided harness, the agent reaches competitive accuracy on challenging benchmarks such as LVBench and VideoMME-long at a smaller input-context cost than packing the full video or performing extensive preprocessing, by adaptively exploring only the evidence needed for the query.
We summarize the three contributions behind this result as follows.
\begin{itemize}

    \item \textbf{Purely online video-agent orchestration and harness.}
    We propose a query-time agent that starts from the raw video file and the user query, decomposes the task into tractable subproblems, invokes expert tools on demand, and aggregates multimodal evidence into a final answer.
    Unlike pipelines that first parse the full video offline or pre-build a hierarchical content database index, the agent acquires evidence only when it is needed for the current query.
    We follow principled agent-design practices and further specify tool functionalities and boundaries to reduce ambiguity and misuse, and use compact VLM tools when the relevant evidence already fits a short-video context.

    \item \textbf{Atomic capability mining and expert-tool design.}
    We mine a taxonomy of fine-grained atomic capabilities from MINERVA expert reasoning traces, and use it as the instrution for tool design to cover various and long-tail cases.
    This yields 60+ expert tools backed by scripts, VLMs, and specialist models (Grounding-DINO~\cite{groundingdino:liu2024grounding}, WhisperX~\cite{whisperx:bain2023whisperx}, PaddleOCR~\cite{cui2025paddleocr30technicalreport}, etc.) for subproblems such as overview, detection, OCR, ASR, and face recognition.

    \item \textbf{Frontier-level accuracy under a compact online budget.}
    On Video-MME-Long, LVBench, and MINERVA, VideoXAgent reaches $85.1\%$, $76.5\%$, and $65.7\%$ with about 50k agent context, even on hour-long videos.
    This is competitive with frontier LMMs such as Seed-2.0-pro~\cite{seed:team2026seed20} and Gemini-3.1-Pro~\cite{gemini31pro:modelcard2026} at a smaller visual footprint than prior video agents.
    The same harness also brings a visually weak or text-only orchestrator to this level, showing that long video performance can come from agentic tool use.
\end{itemize}

\section{Related Work}

\subsection{Large Multimodal Models for Long Video Understanding}

Large Multimodal Models (LMMs) have shown native capabilities in handling long video understanding tasks due to their long-context capabilities, which can put all multi-frame visual tokens into the context window to comprehensively capture the video content.
Powerful open-source and commercial LMMs such as Qwen-VL series models~\cite{qwenvl:bai2025qwen25vl,qwenvl:bai2025qwen3vl}, Kimi-K2.5/K2.6~\cite{kimi:team2026k25}, Seed model series (Seed1.5-VL/1.8/2.0)~\cite{seedvl:team2025seed15vl,seed:team2026seed18,seed:team2026seed20}, and Gemini~2.5/3/3.1 pro~\cite{gemini:team2024gemini15,gemini25pro:modelcard2025,gemini3pro:modelcard2025,gemini31pro:modelcard2026} have shown strong performance on long video understanding benchmarks.
However, as video duration grows, their performance remains tightly coupled to the frame-sampling strategy and to the model's long-context capabilities.

\subsection{Long Video Representation and Retrieval}

Long video methods commonly reduce the raw visual stream into a searchable intermediate representation before answering a query.
VideoRAG constructs graph-based textual knowledge together with multimodal context for retrieval over extremely long videos~\cite{videorag:ren2025retrieval}.
DrVideo converts long video understanding into a document-retrieval problem: it first transforms the video into a coarse text document, retrieves question-related frames, and then uses an agent loop to iteratively augment the document until enough evidence is available to answer~\cite{drvideo:ma2025document}.
VideoTree builds a query-adaptive hierarchical tree of keyframes and aggregates multi-granularity evidence in a coarse-to-fine manner for LLM reasoning~\cite{videotree:wang2024adaptive}.
MR.~Video instead applies a MapReduce principle: it densely analyzes short clips in parallel (Map) and then aggregates clip-level evidence for global reasoning (Reduce)~\cite{mrvideo:pang2025mapreduce}.
DVD builds a multi-granular database of segmented clips and exposes search-centric tools over that database~\cite{dvd:zhang2026deep}, while HAVEN organizes audiovisual entities into a hierarchical tree for agentic search~\cite{haven:yin2026hierarchical}.
These approaches make extended videos tractable by shifting substantial perception and organization into an indexing or preprocessing stage.
Our work instead targets a purely online setting in which evidence is acquired on demand from the raw video for the current query, without requiring a pre-built video database or hierarchical memory.

\subsection{Tool-Augmented Video Agents}

VideoAgent establishes an iterative pattern in which an LLM plans, retrieves a small set of relevant frames through visual tools, and accumulates observations before answering~\cite{videoagent:wang2024long}.
Video-Thinker instead trains an LMM with reinforcement learning to perform temporal grounding and captioning inside the reasoning trace, so that tool-like evidence acquisition is internalized rather than invoked through an external tool interface~\cite{videothinker:wang2025sparking}.
More recent methods strengthen this interaction through different forms of visual tool use within a ReAct loop~\cite{react:yao2023synergizing}: LongVT interleaves reasoning with native temporal cropping and fine-grained resampling~\cite{longvt:yang2026thinking}.
VideoSeek uses a think--act--observe loop and multi-granular tools to seek answer-critical evidence~\cite{videoseek:lin2026long}, DrVideo iteratively retrieves and augments a text document of the video~\cite{drvideo:ma2025document}, and DVD and HAVEN couple agentic search with structured video representations~\cite{dvd:zhang2026deep,haven:yin2026hierarchical}.
Complementary to these video-agent systems, EDC uses off-the-shelf visual specialists to enrich descriptive captions with fine-grained object attributes and relations, demonstrating that specialist evidence can improve multimodal perception and downstream reasoning~\cite{edc:sun2026}.
EDC focuses on image-caption enhancement, sharing the spirit of using visual models as tools, whereas our work orchestrates heterogeneous expert tools for query-conditioned, purely online evidence acquisition over long videos.
Together, these works show the value of adaptive evidence acquisition, but they primarily specify their tool sets through system-level design choices or bind agent actions to a particular indexed representation.
VideoXAgent instead derives an atomic capability taxonomy from expert reasoning traces and maps those capabilities to heterogeneous expert tools, so that a query-conditioned agent can plan and decompose the task without a pre-built long video representation.

\section{Method}

\subsection{Overview}

\begin{figure}[t]
\centering
\includegraphics[width=\textwidth]{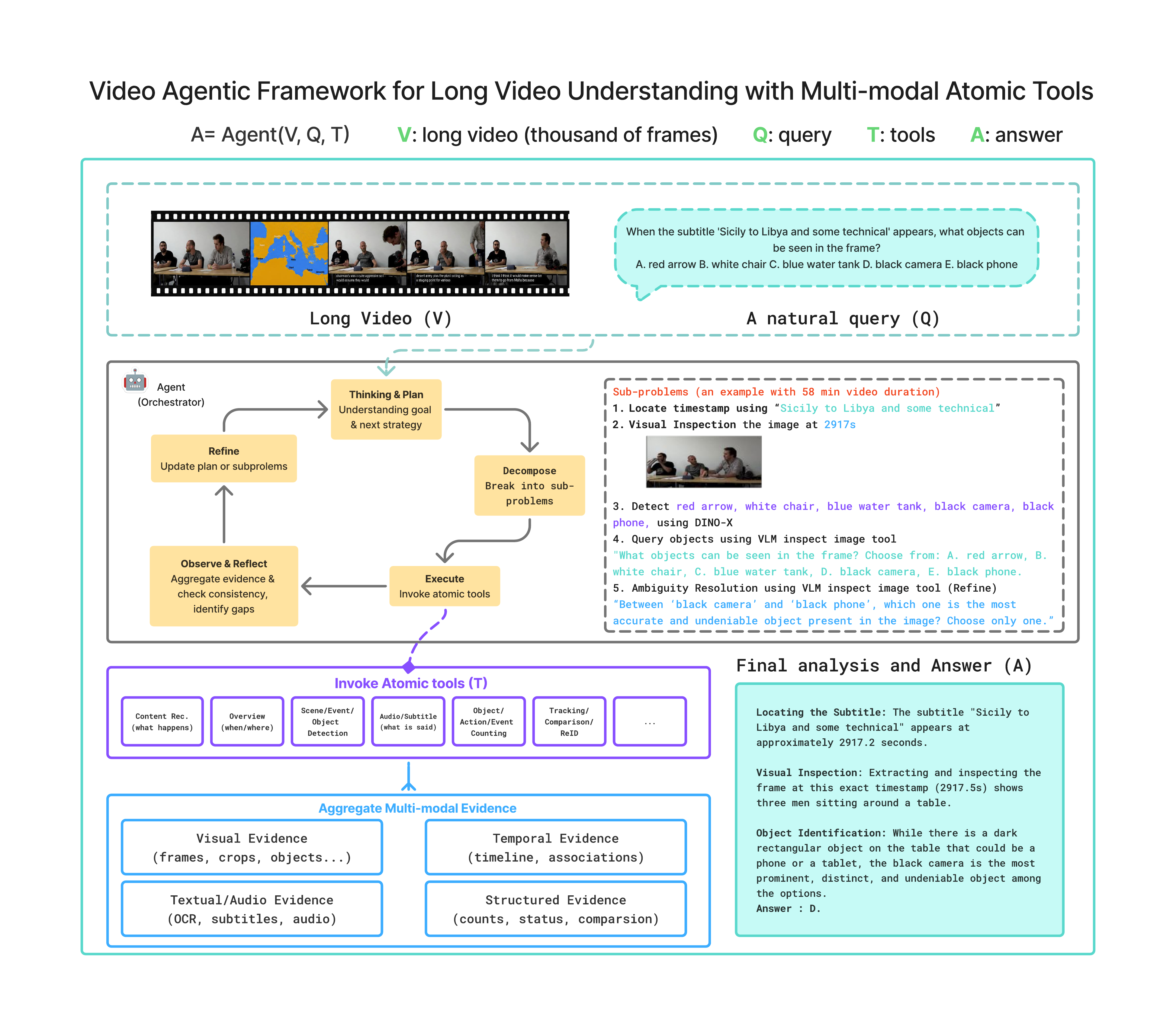}
\caption{System architecture overview. At inference time, given a long video (e.g., 3600 frames, up to several million tokens if being processed by a VLM), the agent decomposes user queries into sub-problems, invokes expert tools, and aggregates multi-modal evidence.}
\label{fig:architecture}
\end{figure}

\subsubsection{Problem formulation}
\label{sec:problem_formulation}

Given a video file $V$ and a natural-language query $Q$, the goal of the video-understanding task is to produce an accurate answer $A$ that is consistent with the video content and directly addresses the query.
Our setting is purely online: the agent starts from the raw input video at query time and may inspect or transform only the portions required by the current reasoning trajectory, rather than assuming an offline preprocessing pass or a preconstructed hierarchical representation of the full video.
We formulate this process as $A=\operatorname{Agent}(V,Q,T,B)$, where $T=\{t_1,\ldots,t_M\}$ denotes the set of expert tools available to the agent and $B$ denotes the execution budget.
Here, $V$ refers to the path of the video file rather than a sequence of decoded video frames placed directly in the model context.
At step $k$, the agent maintains a state $s_k=(Q,H_k,b_k)$, where $H_k$ contains the accumulated tool calls, observations, and evidence, and $b_k$ is the remaining budget.
Based on this state, the agent either invokes a tool with query-specific arguments or produces the final answer.
The sequence of states, tool calls, and observations forms an execution trajectory $\tau$.

Our approach consists of three components:
(1) A data-driven capability taxonomy that provides principled guidance for tool design.
(2) A suite of 60+ expert tools and compositional workflows built upon this taxonomy. Each tool targets a specific atomic sub-problem. The workflows chain these tools for high-level reasoning.
(3) A systematic video-agent harness that orchestrates these tools at inference time by adapting to the query and video duration, decomposing the query into sub-problems, invoking appropriate tools, aggregating multimodal evidence, and controlling the execution budget to prevent unbounded loops.

Overall, the agent operates in an iterative loop, as illustrated in Figure~\ref{fig:architecture}: \textit{thinking and planning}, \textit{decomposition}, \textit{execution} through tool invocation, \textit{reflection}, and \textit{refinement}.
The loop continues until the agent obtains sufficient evidence to produce an answer or the remaining budget cannot support another tool round, at which point it returns an answer using the evidence already collected.

\subsection{Capability Mining and Expert Tool Design}

\begin{figure}[t]
\centering
\includegraphics[width=\textwidth,height=0.36\textheight,keepaspectratio]{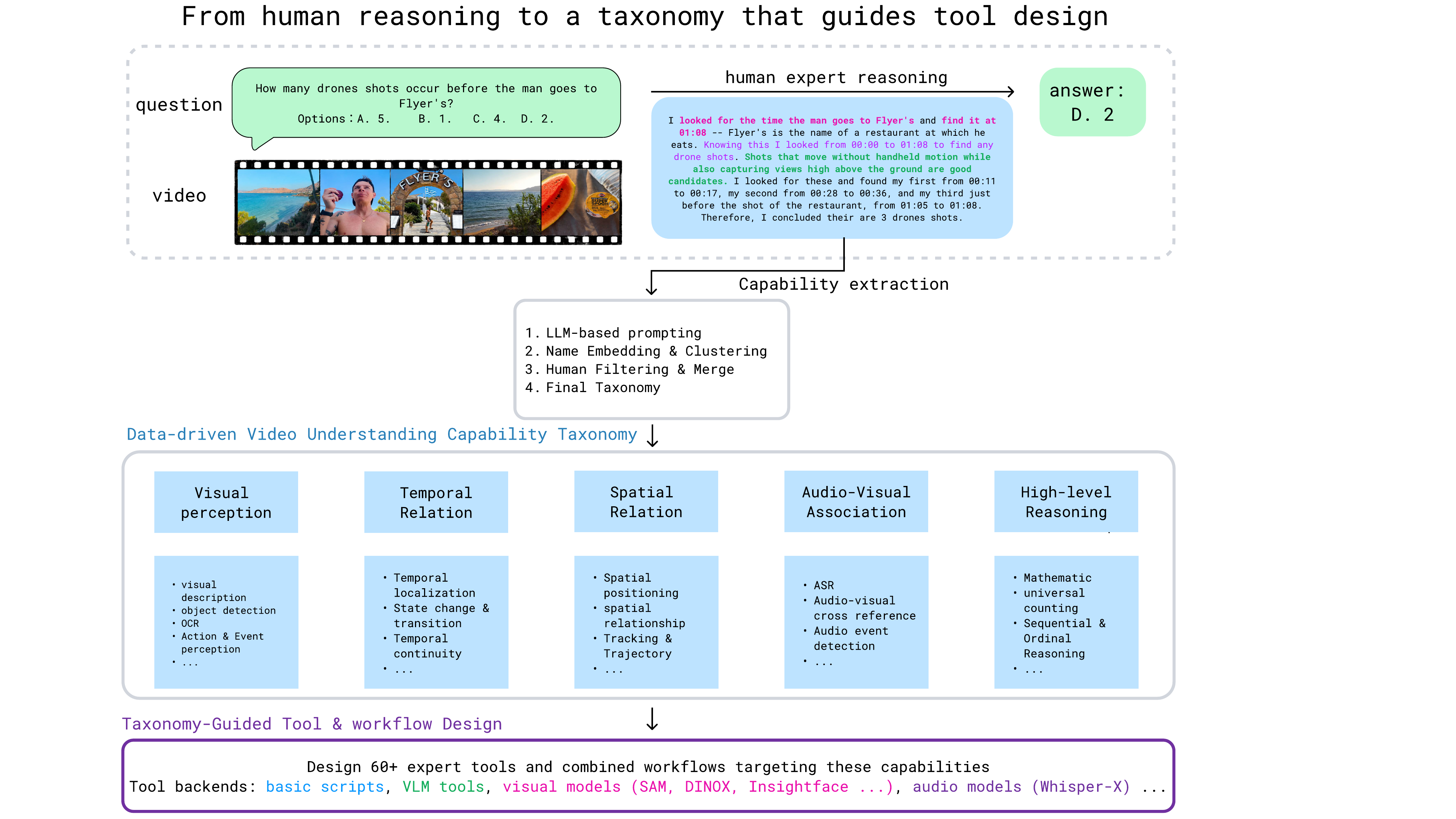}
\caption{Capability mining pipeline. Starting from MINERVA human-expert reasoning traces $(Q,R,A,O)$, we extract atomic capabilities by LLM prompting, merging clustering, and refine the result with human filtering. The resulting five-category taxonomy then guides the design of expert tools and compositional workflows.}
\label{fig:capability_mining}
\end{figure}

Existing benchmarks such as LongVideoBench~\cite{longvideobench:wu2024}, VideoMME\cite{videomme:fu2025}, and LVBench~\cite{lvbench:wang2025} mainly classify question formats, video genres, or coarse task types; they therefore provide limited guidance for decomposing complex tasks into atomic capabilities and equipping models or agents with tools that cover them.
To address this gap, rather than subjectively defining the required capabilities, we take a data-driven approach and mine a taxonomy of atomic capabilities from human reasoning traces.
Figure~\ref{fig:capability_mining} summarizes this pipeline.

Specifically, we use the MINERVA~\cite{minerva:nagrani2025evaluating} benchmark as our primary source, because it provides detailed human-expert reasoning annotations as $(Q, R, A, O)$ tuples, where $Q$ denotes the question, $R$ the human reasoning trace, $A$ the ground-truth answer, and $O$ the candidate options when available.
By analyzing these reasoning traces, we observe the cognitive trajectories that human experts follow when solving complex video understanding questions, thereby inspiring the design of various expert tools that video agents can use.

It must be explicitly emphasized that the only information extracted from these annotations is a high-level capability taxonomy for general video understanding, rather than any ground-truth instance data.
We adopt this extracted taxonomy purely as a design blueprint to construct dedicated expert tools and orchestrate their compositional workflows for tackling complex video tasks.
No concrete example data is exposed, memorized, or disclosed.

Based on this principle, our bottom-up mining process consists of four main steps:
\begin{enumerate}
    \item \textbf{LLM Prompting:} For each task quadruple $(Q, R, A, O)$ (representing the query, human reasoning trace, ground-truth answer, and candidate options), we prompt an LLM to reverse-engineer the cognitive trajectory $R$. To prevent hallucination, the model must output a structured JSON schema where each extracted capability is explicitly grounded by quoting direct \textit{reasoning\_evidence} from $R$.
    \item \textbf{Merging \& Filtering:} We first normalize surface-form variants, then merge long-tail, low-frequency capabilities into semantically similar high-frequency ones using cosine similarity on \texttt{all-MiniLM-L6-v2} embeddings~\cite{minilm:wang2020minilm,sbert:reimers2019sentencebert}.
    \item \textbf{Embedding \& Clustering:} We encode the merged capability names into dense vectors and apply $K$-means clustering, selecting $K$ by the silhouette score, then we manually review to refine cluster boundaries.
    \item \textbf{Final Taxonomy:} This iterative process yields a consolidated taxonomy of 22 atomic capabilities across 5 distinct categories: \textit{Visual Perception, Temporal Relation, Spatial Relation, Audio-Visual}, and \textit{High-Level Reasoning}.
\end{enumerate}
Details of the mining pipeline are given in the Appendix (Sec.~\ref{sec:mining_procedure}).

A key insight derived from this methodology is that essential capabilities must be discovered from the actual reasoning process of solving video questions, rather than inferred from macroscopic task-level labels or subjective assumptions.

An illustrative decomposition of a paraphrased sports query is provided in Appendix Table~\ref{tab:example_qa}. It shows how a query that appears to require only a coarse task label can involve several fine-grained capabilities in the underlying reasoning trace.
As detailed in Table~\ref{tab:capability_taxonomy}, our framework uses these trace-level observations to construct a bottom-up taxonomy rather than relying on top-down task labels.
By grounding tool design in observed human reasoning trajectories, we aim to equip our agent with atomic capabilities relevant to complex video reasoning.

The mined taxonomy guides both atomic tool design and workflow composition.
Well-defined perceptual or retrieval capabilities map to focused tools, whereas comparison, cross-modal fusion, and logical inference map to multi-tool workflows, as summarized in Table~\ref{tab:mapping}.

Each tool specifies its media scope, task-specific inputs, and expected outputs, while hiding backend-specific execution behind a common interface (Appendix Table~\ref{tab:tool_specs}).
Tool observations are aligned by video, temporal, and spatial references, enabling workflows such as temporal localization followed by dense inspection and targeted OCR, detection, or audio analysis.
Empty results and execution errors remain explicit observations, allowing the agent to revise the call, select another tool, or answer from the available evidence.

\subsection{Video Agent Orchestration and Harness}

\subsubsection{ReAct-style Agent Loop}
\label{sec:react_loop}

Overall, our video agent follows a ReAct-style paradigm consisting of the following steps:
\begin{enumerate}
    \item \textbf{Think and Plan:} At the outset, the agent reasons solely from the user query and identifies the required information.
    The LLM initially generates a rough plan based on its understanding of the question and the provided options.
    Since the agent has no prior knowledge of the video content, it first uses basic tools to obtain the video duration and may invoke audio or subtitle tools to acquire preliminary cues.
    Once a coarse picture of the video emerges, the agent decomposes the user query into multiple subproblems and records their progress in the execution trajectory.
    Note that the plan may also be updated or revised based on observations from tool outputs.
    Some overview or scene caption tools may also be invoked to build a high-level understanding of the video.
    \item \textbf{Decompose and Act:} For each subproblem, the agent maps the required capability to an atomic tool or compositional workflow, as summarized in Table~\ref{tab:mapping}.
    It then instantiates task-specific arguments, such as a subquery, sampled frames, a spatial region, or start and end timestamps.
    The selected backend analyzes this bounded media scope and returns an observation for the next reasoning step.
    Section~\ref{sec:tool_selection} further details how tools and workflows are selected and how their observations are aggregated.
    \item \textbf{Observe and Reflect:} The agent links the observation to its tool call and media scope, then determines whether it resolves the current subproblem or reveals missing, ambiguous, conflicting, or failed evidence.
    \item \textbf{Continue or Finish:} If required evidence remains unresolved, the agent updates the plan and continues the loop.
    Otherwise, or when another useful round exceeds the remaining budget, it stops tool use and synthesizes the final prediction.
\end{enumerate}
The orchestrator is instructed by a single system prompt that encodes this loop together with tool-efficiency limits, duration-aware routing, and a required answer format.
The full prompt is given in Appendix~\ref{app:system_prompt}.

\subsubsection{Tool/Workflow Selection and Evidence Aggregation}
\label{sec:tool_selection}

\paragraph{Tool and workflow selection.}
Tool selection is conditioned on the current subproblem, the required modality, and the media scope already identified in the trajectory.
The agent uses direct VLM inspection for general visual interpretation and routes specialized requirements to the corresponding expert tools, such as OCR for embedded text, transcription for speech, detection or tracking for object-centric questions, and temporal retrieval for event localization.
When a subproblem requires several dependent operations, the agent invokes a workflow whose outputs narrow the inputs of subsequent tools.
Independent analyses over multiple segments or modalities can be batched or executed in parallel when their interfaces permit, while dependent calls remain sequential.

\paragraph{Objective visual evidence prompting.}
Because the agent decomposes the user query into intermediate VLM sub-queries, these sub-queries may contain implicit presuppositions that take the existence of an event, object, action, or temporal relation as given before any visual check.
A VLM may then confirm such assumptions even when the supplied visual evidence is incomplete or ambiguous.
To reduce this failure mode, we wrap every VLM-based tool with a shared \emph{objective-evidence} contract that treats the incoming sub-query as a hypothesis to be verified rather than a fact to be accepted.
Under this contract, the model grounds its response only in the media supplied to the current call, such as sampled frames, image crops, thumbnail grids, or scene segments.
Auxiliary signals, including subtitles, OCR strings, and detector outputs, may be used as tool-generated context, but must be explicitly separated from directly visible evidence.
The model is also instructed not to assert identity, intent, emotion, causality, exact counts, small text, or fine-grained spatial relations unless the supplied evidence is sufficiently clear; otherwise, it should state the limitation explicitly.
Each VLM tool returns a structured response with \textsc{visual\_evidence}, \textsc{answer}, and \textsc{uncertainty} fields, with tool-specific variants for scene captioning and directed scene queries.
This structure allows the orchestrator to distinguish grounded observations from interpretations and to propagate uncertainty during multi-tool evidence aggregation.
We refer to these constraints as the \emph{objective rules}; their contribution is evaluated in Table~\ref{tab:ablation_study}.

\paragraph{Evidence aggregation and conflict resolution.}
Each observation is associated with its originating tool and available temporal or spatial reference before being added to the evidence state.
The agent aggregates complementary visual, textual, audio, and structured observations according to their relevance and specificity to the query rather than assuming that all outputs are equally reliable.
When observations conflict, it first checks whether they refer to different times, regions, entities, or interpretations; unresolved conflicts trigger targeted verification with refined inputs or an alternative tool.
Final answer synthesis uses only the evidence retained after conflicts are resolved, distinguishes direct observations from agent inference, and answers the query or selects an option without reproducing the full execution trace.

\paragraph{Illustrative example.}
To answer a sports query about which play sequence caused the score to become $24$--$13$, the agent first uses temporal retrieval or scene captions to identify candidate scoring intervals.
Within those intervals, OCR traces scoreboard changes, while visual inspection or action recognition identifies the corresponding plays.
The observations are aligned by timestamp to construct the event sequence; if a recognized play does not match the scoreboard transition, the agent resamples that interval more densely or verifies it with an alternative tool before answering.

\subsubsection{Adaptive Strategy for Video Length}
\label{sec:adaptive_length}

\paragraph{Routing principle.}
Video duration is an important routing signal, but the effective path also depends on whether the query-relevant visual and audio evidence can fit within the available model context.
The agent therefore obtains basic video metadata first and adapts the temporal coverage and sampling granularity during execution rather than relying on a fixed duration threshold.

\paragraph{Short-video path.}
When the relevant content can be examined within a compact context, the agent favors direct video or multi-frame VLM analysis to preserve global event continuity.
Expert tools are invoked selectively when the query requires information that broad visual inspection does not reliably expose, such as small text, a specific object or face, precise speech content, or a localized temporal relation.
This path avoids constructing an unnecessary hierarchy and stops once the direct analysis and any targeted verification provide sufficient evidence.

\paragraph{Long-video path.}
For videos whose relevant evidence cannot be covered compactly, the agent follows a coarse-to-fine exploration strategy.
It first builds a sparse temporal map from overview, scene, subtitle, or audio observations selected according to the query.
It then localizes candidate intervals and applies denser visual sampling or specialized tools only within those intervals.
If the evidence remains incomplete or inconsistent, the agent expands adjacent intervals, increases sampling density, or switches modality; it does not repeatedly resample the full video.
The process terminates early when the required subproblems are resolved, while the shared execution budget and forced-answer mechanism bound unsuccessful exploration.

\paragraph{Budget control.}
\label{sec:budget_warning}
A failure mode of ReAct-style agents is \emph{non-termination}: when poor planning yields tool calls that return no useful evidence, the agent keeps repeating nearly the same calls; the many low-value observations that follow consume and pollute the context, which further degrades planning and traps the agent in a tool-calling loop that ends with \emph{no} answer---wasting the entire trajectory's compute. We control this with two complementary mechanisms, implemented as a lightweight middleware that operates purely on the message stream.
\emph{(i) Hard step limit with forced answering.} Each trajectory is capped at $L{=}125$ node executions, counting model steps, tool-call messages, and corresponding tool-execution steps. Under the common three-node pattern, this is approximately 42 tool calls. When the remaining budget can no longer support another tool round, the middleware disables all tools (\texttt{tool\_choice=none}) and the final model call must emit a textual prediction---so every run is guaranteed to produce an answer.
\emph{(ii) Progressive budget warnings.} At $L{=}125$, most trajectories finish within $65\%$ of the budget; runs that go beyond are almost always struggling. When consumption crosses $65\%$/$80\%$/$90\%$ of $L$, the middleware therefore injects a single escalating message (\textsc{Notice}/\textsc{Warning}/\textsc{Critical}, each fired once) that reports the remaining call budget and instructs the agent to stop redundant verification, resolve conflicting evidence by majority vote, and commit to a final answer. Both mechanisms leave the tools and decoding untouched; their effects are quantified in Sec.~\ref{sec:abl_warning}.

\section{Experiments}
\subsection{Experimental Setup}

We evaluate VideoXAgent on five long video understanding benchmarks that span diverse durations, themes, and reasoning demands.
Unless otherwise noted, we report results on the full official evaluation splits (LongVideoBench-long, Video-MME-long, and LVBench); for MINERVA we use the publicly released set, and for Video-MME-v2 a stratified subset, as detailed below.
Table~\ref{tab:sota_comparison} summarizes the main comparison.
\begin{itemize}
    \item \textbf{LongVideoBench}~\cite{longvideobench:wu2024} contains web videos of varying lengths up to one hour, together with subtitles, covering diverse themes and evaluating detailed retrieval and reasoning over long videos.
    We report results on its long split, which comprises 564 questions from 188 videos with durations between 900 and 3600 seconds.
    \item \textbf{Video-MME}~\cite{videomme:fu2025} is a comprehensive multimodal benchmark for long video understanding across diverse video types and temporal ranges.
    We evaluate on its long subset, which includes 900 questions from 300 videos with an average duration of 2{,}466 seconds.
    \item \textbf{LVBench}~\cite{lvbench:wang2025} focuses on long-term memory and extended comprehension over multimodal inputs.
    It consists of 103 publicly sourced videos (e.g., TV series, sports, and surveillance) totaling about 117 hours, with an average duration of 4{,}101 seconds and 1{,}549 human-annotated questions; we evaluate on the full official set.
    As the original LVBench data has no subtitles, we disable transcripts, ASR, and other audio tools and use visual frames only, matching the \textit{wo.\ sub} protocol in Table~\ref{tab:sota_comparison}.
    \item \textbf{MINERVA}~\cite{minerva:nagrani2025evaluating} targets complex, multi-step video reasoning.
    The paper reports about 1.5K hand-crafted questions over videos ranging from under 2 minutes to over 1.5 hours, spanning domains such as short films, sports, and instructional content, with five answer choices and a detailed human reasoning trace per question.
    We evaluate on the publicly released JSON from the Neptune repository,\footnote{\url{https://github.com/google-deepmind/neptune\#minerva}} which contains 1{,}357 questions.
    \item \textbf{Video-MME-v2}~\cite{videommev2:2026} is a newly released progressive benchmark with 800 YouTube videos (average 10.4 minutes; 99\% under 20 minutes) and 3{,}200 eight-way questions, each video bound to a fixed group of four.
    Considering the cost, we evaluate a video-level stratified subset of 118 videos / 472 questions: 6 videos from each of 10 \textit{relevance} \texttt{second\_head} classes, and roughly 3 videos from each of 21 \textit{logic} Q4 \texttt{third\_head} classes, keeping groups intact and category coverage near-uniform.
    The subset balances independently labeled \textit{relevance} groups (60 videos / 240 QA) and causally chained \textit{logic} groups (58 videos / 232 QA).
    This yields exact uniformity over the 10 relevance classes and full coverage of all 21 logic classes.
\end{itemize}
For the LongVideoBench ablation in Table~\ref{tab:ablation_study}, we draw a category-stratified sample of 150 questions from the long split, using the benchmark's question categories as strata.
For the budget-control study in Sec.~\ref{sec:abl_warning}, we evaluate on the 217-question loop-prone stress set defined in Appendix~\ref{app:warn_provenance}.

For VideoMME-v2, we report the benchmark's nonlinear score together with the fine-grained capability breakdowns summarized in Figure~\ref{fig:videomme_v2_ablation}.
The performance of LMMs on Video-MME-Long and Video-MME-v2 in Table~\ref{tab:sota_comparison} is from our own evaluations: we call each model's official API following its published video-understanding configuration, on the same splits as our agent (the official long split for Video-MME-Long; the 118-video / 472-question stratified subset for Video-MME-v2).

In the comparison and ablation tables, the frames/context column reports processed frame counts per-example and agent context lengths when available. For directly evaluated LMMs, frames and context refer to the same visual input consumed by the model, reported as a frame count or a context-token count, respectively. For video agents, the two quantities are distinct: frame counts refer either to frames processed by VLM tools or to frames materialized during offline index construction in two-stage agents such as DVD and HAVEN, while context denotes measured agent maximum context tokens when reported.

We map the final assistant's answer format to the official options; a wrong match or no match is counted as incorrect.

\begin{table*}[!tbp]
    \centering
    \caption{Comparison with SOTA video agents and large multimodal models on long video understanding benchmarks. The frames/context column records the average per-example visual input frames for VLM tools and the maximum prompt context for the agent or LMM (agent context in DVD, HAVEN, and VideoSeek is not reported). LMM results on Video-MME-Long and Video-MME-v2 are from our evaluations under each model's published configuration using official-APIs. Our agent uses Claude-Opus-4.6 with Gemini-3.1-Pro-preview VLM tools. LMM LVBench numbers follow the source default and may include native audio. 
    $^*$This is the result of visual-only ablation (71.7) in the paper of HAVEN. $^\dagger$Seed-2.0-pro does not report per-example context on LVBench; this result is the average input tokens when we use its official API for the LVBench samples. }
    \label{tab:sota_comparison}
    \scriptsize
    \renewcommand{\arraystretch}{1.15}
    \setlength{\tabcolsep}{3pt}
    \resizebox{\textwidth}{!}{%
    \begin{tabular}{cc|lc|lc|lc|ccc}
    \toprule
    \multirow{2}{*}{\textbf{Method}} &
    \multirow{2}{*}{\textbf{Type}} &
    \multicolumn{2}{c}{\makecell{\textbf{VideoMME}\\\textbf{Long} \textit{(w.\ sub)}}} &
    \multicolumn{2}{c}{\makecell{\textbf{LongVideoBench}\\\textbf{Long} \textit{(w.\ sub)}}} &
    \multicolumn{2}{c}{\makecell{\textbf{LVBench}\\\textit{(wo.\ sub)}}} &
    \multicolumn{3}{c}{\makecell{\textbf{VideoMME-v2}\\\textbf{Stratified ($n{=}472$)}\\\textit{(w.\ sub)}}} \\
    \cmidrule(lr){3-4}
    \cmidrule(lr){5-6}
    \cmidrule(lr){7-8}
    \cmidrule(lr){9-11}
    & &
    \textbf{Acc.} & \textbf{frames/Context} &
    \textbf{Acc.} & \textbf{frames/Context} &
    \textbf{Acc.} & \textbf{Context} &
    \textbf{Acc.} &  \textbf{Non-lin Score} &    \textbf{Context} \\
    \midrule
    Doubao-Seed-2.0-pro~\cite{seed:team2026seed20} & LMM & 85.5\% & $\sim$112k & -- & -- & 76.4\% & $\sim$ 96k$^\dagger$ & 57.63\% & 40.26 & 49.1k \\
    Gemini-3.1-Pro-preview~\cite{gemini31pro:modelcard2026} & LMM & 86.2\% & $\sim$179k & -- & -- & -- & -- & 59.96\% & 38.47 & 43.2k \\
    \hline
    DVD~\cite{dvd:zhang2026deep} & Video Agent & -- & -- & 68.6\% & 2,816\,f/- & 74.2\% & 8,074\,f/- & -- & -- & -- \\
    VideoSeek~\cite{videoseek:lin2026long} & Video Agent & 81.2\% & 15.9\,f/- & 73.5\% & 29.6\,f/- & 68.4\% & 92.3\,f/- & -- & -- & -- \\
    HAVEN~\cite{haven:yin2026hierarchical} & Video Agent & 82.8\% & 1,645\,f/- & 78.2\% & 943.5\,f/- & 71.7\%$^*$ & 2,702\,f/- & -- & -- & -- \\
    \hline
    Ours & Video Agent & 85.1\% & 52.0\,f/51k & 78.1\% & 116.7\,f/55.3k & 76.5\% & 60.8\,f/55k & 61.28\% & 40.74 & 53.5k \\
    \bottomrule
    \end{tabular}
    }
\end{table*}

\subsection{Implementation}

We implement VideoXAgent as a ReAct-style orchestrator over a unified tool interface rather than as a monolithic video encoder.
The toolkit is organized around the mined capability taxonomy in Appendix Table~\ref{tab:capability_taxonomy}: each atomic capability is backed by one or more callable tools, while High-Level Reasoning capabilities are realized as compositional ReAct workflows rather than single calls (Appendix Table~\ref{tab:mapping}).
As detailed in Appendix Tables~\ref{tab:tool_specs} and~\ref{tab:backend_technologies}, the current system exposes $60+$ atomic tools spanning $14$ backend families---including VLM-based visual inspection, Grounding-DINO/DINO-X detection~\cite{groundingdino:liu2024grounding,dinox:ren2024dinox}, PaddleOCR~\cite{cui2025paddleocr30technicalreport}, WhisperX~\cite{whisperx:bain2023whisperx}, PyAnnote~\cite{pyannote:bredin2020neural}, SAM/SAM2~\cite{sam:kirillov2023segment,sam2:ravi2025sam2}, CLIP retrieval~\cite{clip:radford2021learning}, FFmpeg-based media extraction, and custom Python utilities---grouped into categories such as detection, OCR, temporal/spatial analysis, audio processing, face \& emotion, and basic video operations.
Each tool receives a bounded media scope together with task-specific arguments such as sampled frames, timestamps, spatial crops, or sub-queries, and returns a structured observation that can be merged into the shared evidence state.
Video preprocessing therefore remains query-conditioned: the system operates on the raw video path, uses FFmpeg or script backends (Appendix Table~\ref{tab:backend_technologies}) only to extract the clips, frames, audio spans, or subtitle-aligned segments required by the current step, and avoids a separate offline full-video parsing stage.
An exception is LVBench, where subtitle, ASR, and other audio tools are disabled and only visual frames are used.
For VLM-based tools, we reuse the objective-evidence prompting protocol of Sec.~\ref{sec:tool_selection}, which records grounded evidence, answers, and uncertainty separately so that later aggregation can distinguish direct observations from inferred conclusions.
The orchestrator itself is conditioned by the system prompt reproduced in Appendix~\ref{app:system_prompt}.

The default execution budget follows Sec.~\ref{sec:budget_warning}: each trajectory is capped at $L{=}125$ node executions, with escalating warnings at $65\%$, $80\%$, and $90\%$ of the budget and forced answering when another tool round no longer fits.
Independent tool calls over disjoint temporal segments or modalities may be batched or run in parallel when their interfaces permit.

We primarily use Claude-Opus-4.6 as the orchestrator, leveraging its strong agentic capabilities for competitive long video understanding.
Its native long video perception remains comparatively limited; the observed gains stem from planning, tool use, and evidence aggregation, rather than from packing dense frames into a single context window.
We further evaluate and ablate alternative orchestrators, including GLM-5.2~\cite{glm:team2026glm5}, Kimi-K2.6~\cite{kimi:team2026k25}, Gemini-3.1-Pro~\cite{gemini31pro:modelcard2026}, DeepSeek-V4-Flash (2026-04-24-version)~\cite{deepseek:team2026v4}, and Doubao-Seed-2.0~\cite{seed:team2026seed20}.
For VLM-based tool backends, we mainly use Gemini-3.1-Pro-preview~\cite{gemini31pro:modelcard2026} via API and Qwen3.6-35B-A3B~\cite{qwen:team2026qwen36} via local deployment.

\subsection{Comparison with State-of-the-Art Large Multimodal Models and Video Agents}

We compare our approach with representative video agents and frontier multimodal models across multiple long video understanding benchmarks. For each benchmark subset, we report accuracy and the number of input frames or maximum context length when available.

\begin{table}[!tbp]
    \centering
    \caption{Comparison on MINERVA against frontier LMMs. For our agent rows, we report the orchestrator (Agent) and VLM tool backend; LMM baselines do not use an external tool stack. Context is mean (min$\sim$max) tokens per example; Agent Steps is the mean number of tool-loop steps. $^\dagger$Result from the original MINERVA paper~\cite{minerva:nagrani2025evaluating} (1{,}515 questions, with ASR; context tokens converted from the reported frames at 256 tokens/frame). $^\ddagger$Gemini-3-Pro-preview result reported in the Doubao-Seed-1.8 paper~\cite{seed:team2026seed18}.}
    \label{tab:minerva_comparison}
    \footnotesize
    \renewcommand{\arraystretch}{1.12}
    \setlength{\tabcolsep}{3pt}
    \sbox{\minervabox}{%
    \begin{tabular}{@{}l|ccc|lcc@{}}
    \toprule
    \textbf{Method} & \textbf{Type} & \textbf{Agent} & \textbf{VLM} & \textbf{Acc.} & \makecell{\textbf{Context}\\{\scriptsize avg (min$\sim$max)}} & \makecell{\textbf{Steps}\\{\scriptsize avg}} \\
    \midrule
    Gemini-2.5-Pro-Thinking~\cite{gemini25pro:modelcard2025} & LMM & -- & -- & 64.7$^\dagger$ & 65.5k (256\,f) & -- \\
    Gemini-2.5-Pro-Thinking~\cite{gemini25pro:modelcard2025} & LMM & -- & -- & 66.2$^\dagger$ & 262.1k (1{,}024\,f) & -- \\
    Gemini-3-Pro-preview~\cite{gemini3pro:modelcard2025} & LMM & -- & -- & 65.0$^\ddagger$ & -- & -- \\
    Doubao-Seed-1.8~\cite{seed:team2026seed18} & LMM & -- & -- & 62.4 & -- & -- \\
    Doubao-Seed-2.0-pro~\cite{seed:team2026seed20} & LMM & -- & -- & 66.5 & -- & -- \\
    \hline
    Ours & Video Agent & GLM-5.2 & Qwen3.6-35B-A3B & 65.4 & 33.7k (17.7k$\sim$189.9k) & 35.6 \\
    Ours & Video Agent & Claude-Opus-4.6 & Qwen3.6-35B-A3B & 65.7 & 37.1k (25.0k$\sim$97.6k) & 27.5 \\
    \bottomrule
    \end{tabular}%
    }%
    \ifdim\wd\minervabox>\columnwidth
      \resizebox{\columnwidth}{!}{\usebox{\minervabox}}%
    \else
      \usebox{\minervabox}%
    \fi
\end{table}

Table~\ref{tab:sota_comparison} shows that VideoXAgent is competitive with frontier long-context LMMs while using a much smaller visual input footprint.
On VideoMME-Long, our agent reaches $85.1\%$ accuracy with $52.0$ frames / $\sim$51k context, close to the strongest direct LMM rows ($85.5\%$--$86.2\%$) while remaining far below the dense-frame budgets reported for prior agents such as HAVEN.
The same pattern appears on LongVideoBench-Long and LVBench, where VideoXAgent improves over DVD and remains close to or above the previously reported agent baselines without requiring uniformly large frame counts.
Video-MME-v2 in this table is a stratified subset ($n{=}472$); both our agent and the LMM baselines in that column are evaluated on this same split, so those numbers are for positioning rather than an official full-split leaderboard.

We further evaluate on MINERVA, a more challenging benchmark for complex video reasoning (Table~\ref{tab:minerva_comparison}; \(n{=}1{,}357\) from the public Neptune release).
On this benchmark, we compare with frontier LMMs including Doubao-Seed-1.8/2.0-pro~\cite{seed:team2026seed18,seed:team2026seed20}, Gemini-2.5-Pro-Thinking~\cite{gemini25pro:modelcard2025,minerva:nagrani2025evaluating}, and Gemini-3-Pro-preview~\cite{gemini3pro:modelcard2025,seed:team2026seed18}.
We adopt two orchestrators with limited native vision: Claude-Opus-4.6, which is strong at coding but comparatively weak visually, and GLM-5.2~\cite{glm:team2026glm5}, a text-only LLM with no visual ability.
With Qwen3.6-35B-A3B~\cite{qwen:team2026qwen36} as the VLM tool backend, both reach frontier LMM level ($65.7\%$ / $65.4\%$ vs.\ $62.4$--$66.5\%$).
This suggests that our harness converts long video understanding into an agentic capability of invoking video-understanding tools, rather than requiring a single model to ingest the full visual context.

\subsection{Ablation study}

\begin{figure*}[t]
\centering
\includegraphics[width=0.72\textwidth]{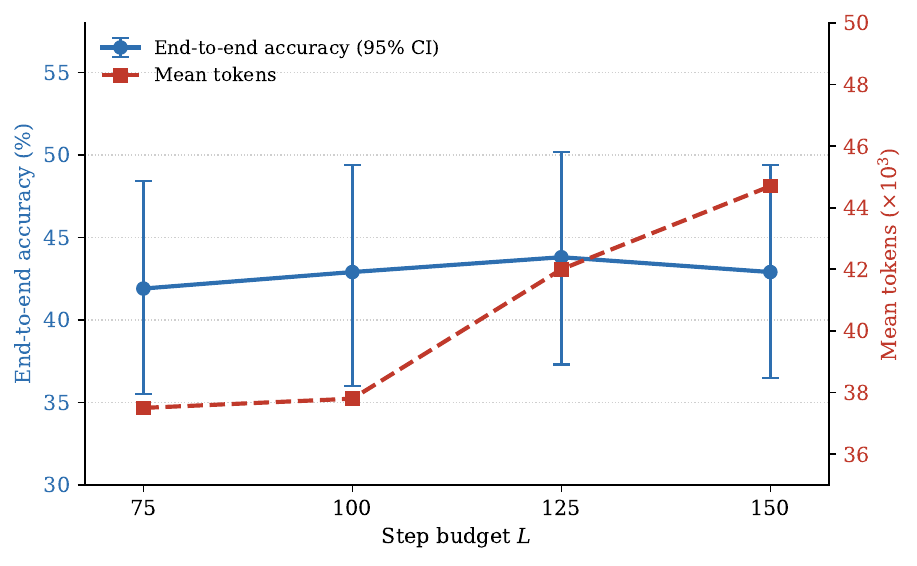}
\caption{Sensitivity to the step budget $L$ on the targeted stress set ($n{=}217$, paired).
The plot shows end-to-end accuracy with $95\%$ CIs and mean token cost, matching the values in Table~\ref{tab:warn_limit}; non-termination is $0\%$ at every limit.}
\label{fig:budget_warning}
\end{figure*}

\begin{table*}[!ht]
    \centering
   
    \caption{Ablation of tool configuration, model choice, and objective rules on a 150-question LongVideoBench-Long sample. Each cell reports accuracy and total input context when available. $^\dagger$~denotes thinking/reasoning mode. This table uses Kimi-K2.6 with Qwen3.6 tools, not the headline Claude-Opus-4.6 + Gemini-3.1-Pro-preview configuration of Table~\ref{tab:sota_comparison}.}

    \label{tab:ablation_study}
    \scriptsize
    \renewcommand{\arraystretch}{1.15}
    \setlength{\tabcolsep}{2.5pt}
    \resizebox{\textwidth}{!}{%
    \begin{tabular}{ccccc|cc|cccc}
    \toprule
    \makecell{\textbf{Sequence-level}\\\textbf{VLM Tool}} &
    \makecell{\textbf{Single-Image}\\\textbf{VLM Tool}} &
    \makecell{\textbf{Visual}\\\textbf{Expert Tools}} &
    \makecell{\textbf{Audio/}\\\textbf{Subtitle Tools}} &
    \makecell{\textbf{Objective}\\\textbf{Rules}} &
    \makecell{\textbf{Agent}\\\textbf{Model}} &
    \makecell{\textbf{VLM Tool}\\\textbf{Model}} &
      \multicolumn{2}{c}{\makecell{\textbf{LongVideoBench}\\\textbf{Long-subset}}} \\
      & & & & & & & \textbf{Acc} & \textbf{Context} \\
      \midrule
      $\checkmark$ & $\checkmark$ & $\checkmark$ & $\checkmark$ & $\checkmark$ & kimi-k2.6 & qwen-3.6$^\dagger$ & 75.33\%  & 32.2k \\
      $\checkmark$ & $\checkmark$ & $\checkmark$ & $\checkmark$ & $\checkmark$ & kimi-k2.6 & qwen-3.6 & 74.67\%  & 30.9k \\
      $\checkmark$ & $\checkmark$ & $\checkmark$ & $\checkmark$ & $\checkmark$ & gemini-3.1-pro & qwen-3.6 & 80.67\%  & 50.1K \\
      $\checkmark$ & $\checkmark$ & $\checkmark$ & $\checkmark$ & $\checkmark$ & kimi-k2.6 & doubao-seed & 74.67\%  & 29.4k \\
      $\checkmark$ & $\checkmark$ & $\checkmark$ & $\checkmark$ & $\checkmark$ & gemma4-31B-dense & qwen-3.6 & 69.33\%  & 26.8k \\
      $\checkmark$ & $\checkmark$ & $\checkmark$ & $\checkmark$ & $\checkmark$ & deepseek-v4-flash & qwen-3.6 & 71.33\%  & 42.2K \\
  \hline
      \xmark & $\checkmark$ & $\checkmark$ & $\checkmark$ & $\checkmark$ & kimi-k2.6 & qwen-3.6$^\dagger$ & 68.67\% & 35.3k \\
      $\checkmark$ & \xmark  & $\checkmark$& $\checkmark$ & $\checkmark$ & kimi-k2.6 & qwen-3.6$^\dagger$ & 70.00\% & 30.4k \\
      $\checkmark$ & $\checkmark$ & \xmark & $\checkmark$ & $\checkmark$ & kimi-k2.6 & qwen-3.6$^\dagger$ & 72.00\% & 25.4k \\
      $\checkmark$ & $\checkmark$ & $\checkmark$ & \xmark & $\checkmark$ & kimi-k2.6 & qwen-3.6$^\dagger$ & 68.67\% & 37.4k \\
      $\checkmark$ & $\checkmark$ & $\checkmark$ & $\checkmark$ & \xmark & kimi-k2.6 & qwen-3.6$^\dagger$ & 72.60\% & 31.6k \\

    \bottomrule
    \end{tabular}%
    }
    \end{table*}

We conduct ablation studies to analyze how different tool groups, rule constraints, agent backbone models, and VLM tool models affect both answer accuracy and input-context cost. The evaluated components include single-image VLM understanding tools, visual expert tools such as detection, segmentation, tracking, and OCR, audio/subtitle tools, and objective rule constraints.

\paragraph{Effect of tool groups.}
The lower block of Table~\ref{tab:ablation_study} shows that the full tool stack is useful because the failure modes are complementary rather than redundant.
Removing the sequence-level VLM tool lowers LongVideoBench-Long accuracy from $75.33\%$ to $68.67\%$, and removing the audio/subtitle tools yields the same score, indicating that long-range temporal continuity and speech-derived evidence are both high-impact components on this sample.
Dropping the single-image VLM tools also causes a clear degradation to $70.00\%$, which is consistent with their role in inspecting localized frames after coarse temporal routing.
Removing the visual expert tools reduces accuracy by $3.33$ percentage points ($75.33\%\rightarrow72.00\%$). This setting also reduces the context budget, falling from $32.2$k to $25.4$k tokens.
These results suggest that specialist visual tools provide targeted evidence rather than merely adding computation: removing them reduces context usage but also lowers accuracy.
Overall, the table supports a layered design in which sequence-level inspection, localized frame analysis, structured visual experts, and audio/subtitle evidence each recover different parts of the long video reasoning problem.

\paragraph{Effect of agent and tool models.}
The upper block of Table~\ref{tab:ablation_study} shows that model selection changes both the accuracy frontier and the context budget.
Among the reported LongVideoBench-Long runs, the Gemini agent with Qwen-3.6 tools attains the highest accuracy ($80.67\%$), but it also uses the largest reported context budget in this table ($50.1$k), so the gain is not free.
The Kimi-K2.6 configuration with Qwen-3.6 thinking tools---the ablation backbone, distinct from the headline Claude-Opus-4.6 setup---is the strongest balanced point in this table: it trails the best-accuracy row by $5.34$ points while using substantially less context ($32.2$k).
With the Kimi orchestrator and tool groups fixed, the Doubao-Seed tool backend achieves $74.67\%$, compared with $74.67\%$ for the non-thinking Qwen-3.6 configuration.
Likewise, replacing the Kimi orchestrator with Gemma-31B-dense or DeepSeek-V4-Flash~\cite{deepseek:team2026v4} reduces accuracy while shifting the context budget in different directions, so neither weakness can be explained by token usage alone.
We therefore treat the orchestrator and the tool model as a coupled design choice: the ablation does not identify a single universally best backbone, but it does show that accuracy, cost, and evidence quality move together when either component changes.

\paragraph{Effect of the objective rules.}
A subtle failure mode of compositional video agents is \emph{evidence contamination}: an ungrounded intermediate VLM observation can be consumed by later reasoning steps as if it were a verified fact, propagated through aggregation, and potentially used to override more reliable evidence.
The objective rules of Sec.~\ref{sec:tool_selection} are designed to mitigate this problem by acting as a per-observation reliability gate rather than merely a prompt-format constraint.
The final row of Table~\ref{tab:ablation_study} isolates this component by disabling the objective-evidence contract while keeping the orchestrator, VLM tool model, and tool groups unchanged.
Accuracy on LongVideoBench-Long decreases from $75.33\%$ to $72.60\%$, suggesting that the way visual evidence is elicited and recorded has a measurable impact beyond the choice of tools alone.
We observe that, without hypothesis-style framing and explicit uncertainty reporting, VLM tool outputs are more likely to conflate directly visible evidence with auxiliary signals such as subtitles, OCR, or detector outputs, allowing unsupported assumptions to enter the evidence state.
This reliability gain comes with a modest context increase, from $31.6$k to $32.2$k tokens.
The additional cost mainly comes from more structured textual responses and occasional low-confidence verification, rather than dense visual re-sampling.
These results indicate that a small structured cost for evidence integrity can improve long video agent reliability, where a single confident but unsupported observation may derail the final answer.

\begin{figure}[!htbp]
    \centering
    \includegraphics[width=\textwidth]{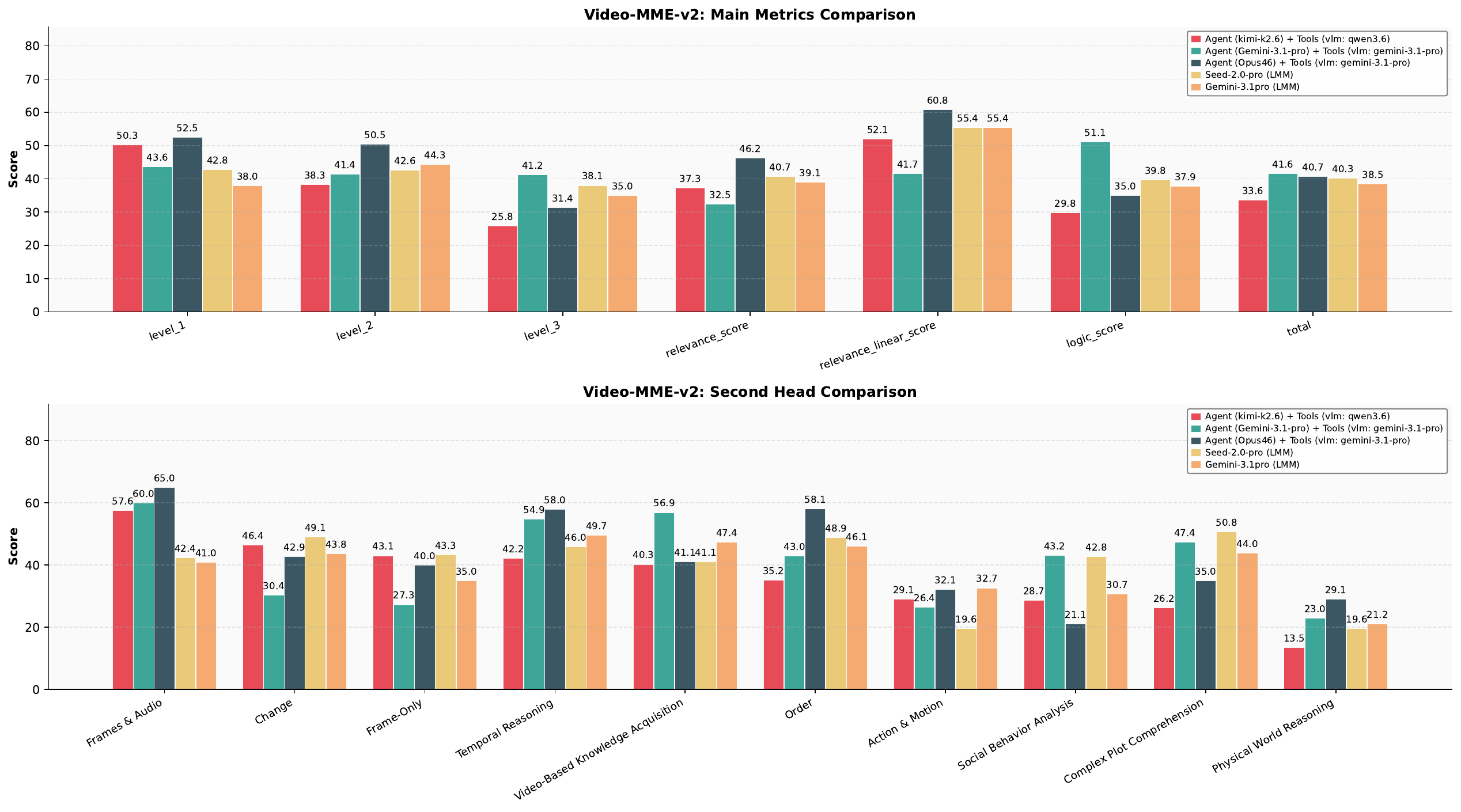}
    \vspace{0.6em}
    \caption{Fine-grained comparison on VideoMME-v2 across different agent backbones, VLM tool models, and direct LMM baselines. The top panel reports main metrics and second-head capability groups, while the bottom panel further breaks down performance by third-head capability categories.}
    \label{fig:videomme_v2_ablation}
\end{figure}

Figure~\ref{fig:videomme_v2_ablation} provides a more detailed view of how model choice and agent design affect VideoMME-v2 performance.
At the aggregate level, the best agent variants and the strongest direct LMM baseline are close but not identical in where they obtain their gains.
The Gemini-3.1-Pro agent paired with Gemini-3.1-Pro tools achieves the highest total score in the plotted set ($41.6$), while the Opus46-based agent is close behind ($40.7$) and exceeds all compared systems on level-1, level-2, relevance-score, and relevance-linear-score metrics.
By contrast, the Gemini-3.1-Pro direct LMM is strongest on logic-score among the plotted configurations, showing that direct long-context reasoning remains competitive on some question types even when the agent variants lead on several retrieval-sensitive metrics.
The second-head breakdown further shows that the strengths are capability-dependent rather than uniform.
The Opus46 agent leads the plotted group on Frames \& Audio, Temporal Reasoning, Order, and Physical World Reasoning, whereas the Gemini-3.1-Pro agent is strongest on Video-Based Knowledge Acquisition and Social Behavior Analysis.
Seed-2.0-pro remains competitive on categories such as Change and Complex Plot Comprehension, so the figure does not support a blanket claim that agentic execution dominates direct LMMs across every capability group.
Instead, the figure supports a narrower conclusion: backbone choice, tool-model choice, and agent orchestration interact with the underlying capability being tested, so they should be selected jointly rather than treated as interchangeable knobs.

\subsubsection{Effect of Budget Control}
\label{sec:abl_warning}

\paragraph{Loop-prone stress set.}
Whether the tool-calling loop of Sec.~\ref{sec:budget_warning} occurs depends
strongly on the orchestrator. With \textsc{Claude-Opus-4.6} we observe no
non-terminating run on any of the five benchmarks, whereas
\textsc{Gemini-3.1-Pro} fails to return a prediction on a subset of questions:
weak task planning leads to tool calls that surface no useful evidence, and
the agent loops until its context is exhausted. We aggregate all such
no-answer cases across VideoMME, VideoMME-v2, MINERVA, LVBench, and
LongVideoBench into a $217$-question stress set---loop-prone \emph{by
construction}, intended as a targeted test of termination behaviour rather
than a random benchmark sample (Appendix~\ref{app:warn_provenance}). Since the
set is selected with a \emph{different} orchestrator than the one evaluated
below, the selection is not circular. All ablations use the
\textsc{Kimi-K2.6} orchestrator with the \textsc{Qwen3.6-A3B} vision expert,
vary only the step limit $L$ and the warning middleware, and share the same
questions (paired design); reported numbers are single runs accompanied by
paired bootstrap $95\%$ confidence intervals (CIs).

\paragraph{Choice of the step limit $L$.}
Table~\ref{tab:warn_limit} sweeps $L\in\{75,100,125,150\}$ with warnings
enabled. The forced-answer mechanism eliminates non-termination at every
limit ($0\%$ throughout), so, on this targeted stress set, the choice of $L$
trades accuracy against cost: accuracy peaks at $L{=}125$ ($43.8\%$) and
saturates---$L{=}150$ is no better
($42.9\%$), and the four $95\%$ CIs overlap---while mean token cost grows monotonically from
$37.5$k to $44.7$k ($+19\%$). We therefore adopt $L{=}125$ ($\approx$42 tool
calls) as the default for all other experiments.

\begin{table}[t]
\centering
\caption{Sweep of the step limit $L$ on the stress set ($n{=}217$, paired,
warnings enabled). Non-termination is $0\%$ at every limit; accuracy peaks at
$L{=}125$, with overlapping $95\%$ CIs across limits, while token cost grows monotonically. $\dag$: default in all other
experiments.}
\label{tab:warn_limit}
\begin{tabular}{lcccc}
\toprule
Limit $L$ & Acc.\ (\%) & $95\%$ CI & Mean tokens & Non-term.\ (\%) \\
\midrule
75            & 41.9          & $[35.5, 48.4]$ & 37.5k & 0.0 \\
100           & 42.9          & $[36.4, 49.3]$ & 37.8k & 0.0 \\
125\,$\dag$   & \textbf{43.8} & $[37.3, 50.2]$ & 42.0k & 0.0 \\
150           & 42.9          & $[36.4, 49.3]$ & 44.7k & 0.0 \\
\bottomrule
\end{tabular}
\end{table}

\paragraph{Effect of the progressive warning.}
Table~\ref{tab:warn_ablation} ablates the warning messages at the fixed
$L{=}125$. Without them, $14.7\%$ ($32/217$) of runs return no answer ($29$
budget exhaustions, $3$ unrecoverable tool errors); with them, such failures
disappear entirely ($0/217$), and
end-to-end accuracy---counting unanswered runs as incorrect---rises from
$35.0\%$ to $43.8\%$ ($+8.8$ points, $95\%$ CI $[+2.3,+15.2]$).
The improvement is largely associated with the $32$ runs that previously returned no answer.
On the remaining $185$ questions, where the baseline already finished, accuracy is $41.1\%$, close to the warning condition's $43.8\%$.
Warnings are injected only after $65\%$ of the budget is used, so trajectories that already finish in time are left unchanged.
Mean token usage simultaneously \emph{drops} by about $12\%$ ($47.5$k
$\rightarrow$ $42.0$k), because looping runs no longer consume the rest of the budget without producing an answer. On this targeted stress set, budget control eliminates the observed non-termination cases while reducing mean token usage.

\begin{table}[t]
\centering
\caption{Progressive budget warning at fixed $L{=}125$ on the stress set
($n{=}217$, paired). End-to-end accuracy counts unanswered runs as incorrect.}
\label{tab:warn_ablation}
\begin{tabular}{lccc}
\toprule
Condition & \makecell{Non-term.\\(\%)} & \makecell{End-to-end\\Acc.\ (\%)} & \makecell{Mean\\tokens} \\
\midrule
w/o warning & 14.7          & 35.0          & 47.5k \\
w/\ warning & \textbf{0.0}  & \textbf{43.8} & \textbf{42.0k} \\
\bottomrule
\end{tabular}
\end{table}

\input{figures/case_study_ora_template.tex}
\chatcasestudyexample

\subsection{Cost and Evidence Efficiency}

We report a narrow but reliable notion of cost: \emph{how much visual and textual context the agent must acquire to answer a question}, and correspondingly \emph{evidence efficiency}: how effectively that context converts into benchmark accuracy.

Under that lens, Table~\ref{tab:sota_comparison} shows that VideoXAgent remains competitive with strong long-context LMMs while operating with selective evidence acquisition instead of uniformly dense video ingestion.
On VideoMME-Long, for example, the agent reaches $85.1\%$ accuracy with $52.0$ sampled frames and roughly $51$k total context, while earlier agent baselines such as DVD and HAVEN report much denser frame usage.
The same selective pattern appears on LongVideoBench-Long and LVBench, where the agent stays within tens to low hundreds of retrieved frames rather than exhaustively expanding the full video into a monolithic context window.

The ablations further show that this cost profile is a property of the full system design rather than of any single model choice.
Removing the visual expert tools lowers the LongVideoBench-Long context budget from $32.2$k to $25.4$k tokens, the largest token saving in the tool-group ablation, but it also reduces accuracy from $75.33\%$ to $72.00\%$.
A lower raw budget is therefore not automatically a better operating point: some apparently expensive tools remain efficient once the quality of the final answer is taken into account.
Sequence-level VLM inspection and audio/subtitle tools produce the largest accuracy drops when removed, even though those ablations do not yield the largest context savings, which implies that these components contribute high-value evidence per unit of added context.

The budget-control study in Table~\ref{tab:warn_ablation} shows that better control can reduce cost and improve accuracy simultaneously: progressive warnings decrease mean token usage from $47.5$k to $42.0$k while eliminating unanswered runs.
Increasing the recursion limit beyond the default range does not improve accuracy on the loop-prone stress set.
Efficiency is therefore improved not only by making each tool call informative, but also by preventing the agent from repeatedly spending context on unproductive branches.
We interpret the agent's main cost advantage as \emph{selective evidence spending}: the system pays for extra tool calls only when they help replace broad, expensive visual inspection with narrower grounded observations.

\subsection{Case Study}

We visualize the case study with a single chat-bubble trace that foregrounds the chronological interaction among the user, the agent, explicit tool calls, and the corresponding tool outputs.
The trace is taken from a real agentic run rather than from a synthetic storyboard, so each step corresponds to an actual tool invocation and returned observation (the description of each step is abstracted for brevity).
It mirrors the method of Secs.~\ref{sec:tool_selection} and~\ref{sec:budget_warning}: the agent first inspects metadata and subtitle availability, then localizes the ``third day of training'' interval with coarse retrieval and scene captioning, extracts subtitle evidence over the candidate window, and finally performs targeted visual verification on the late-Day-3 segment before answering.
This progression shows why the final prediction is not based on a single cue.
The subtitle track surfaces three competing hypotheses---wind, a fall, and a mechanical issue---while the final frame-level inspection resolves the conflict by confirming that the white-top rider is waiting and the green-top rider arrives only after the repair-related dialogue.
The case study therefore illustrates the intended division of labor in VideoXAgent: broad temporal routing narrows the search space, specialized tools recover complementary evidence, and answer synthesis is deferred until conflicting observations are reconciled.
Additional interactive traces, including this example, are available at \url{https://go-agent-x.github.io/video_agent_harness/\#trace}.

\section{Limitations and Conclusion}

We presented VideoXAgent, a purely online video-agent harness that starts from the raw video file and the user query, decomposes the task, invokes expert tools on demand, and aggregates multimodal evidence under a bounded budget.
A data-driven taxonomy of atomic capabilities, mined from MINERVA expert reasoning traces, guides the tool space rather than an ad hoc toolkit.
On Video-MME-Long, LongVideoBench-Long, and LVBench, the agent is competitive with frontier LMMs while using a smaller visual and context footprint.
On the challenging MINERVA benchmark, the same harness brings visually weak or text-only orchestrators to the same range as direct frontier LMMs.
Several limitations remain.
Our efficiency analysis measures the amount of evidence acquired---retrieved frames and the maximum context of the agent model---rather than total token cost.
The capability taxonomy is mined solely from MINERVA reasoning traces and may therefore reflect dataset-specific distributions or omit capabilities that are salient in other domains.
Future work could investigate more compact and efficient tool taxonomies and capability definitions that reduce redundancy while preserving coverage across various domains or adapting to a specific domain.

\bibliographystyle{plain}
\bibliography{refs}

\newpage

\appendix

\newcommand{\appcontentsitem}[3]{%
  \noindent\hyperref[#1]{#2\quad #3\nobreak\leaders\hbox{\normalfont\ .\ }\hfill\pageref*{#1}}\par}
\newcommand{\appcontentssub}[3]{%
  \noindent\hyperref[#1]{\hspace{1.4em}#2\quad #3\nobreak\leaders\hbox{\normalfont\ .\ }\hfill\pageref*{#1}}\par}

\section*{Appendix}
\vspace{-0.3em}
\begingroup
\hypersetup{linkcolor=blue}
\small
\setlength{\parskip}{1.5pt}
\setlength{\parindent}{0pt}
\appcontentsitem{app:capability_taxonomy}{\textbf{A}}{Capability Taxonomy}
\appcontentssub{app:tool_mapping}{A.1}{Detailed Tool Mapping}
\appcontentssub{app:backend_technologies}{A.2}{Backend Technologies}
\appcontentssub{app:tool_specs}{A.3}{Tool Specifications}
\vspace{0.35em}
\appcontentsitem{app:tool_usage}{\textbf{B}}{Tool Usage Distribution Analysis}
\appcontentssub{app:usage_by_category}{B.1}{Usage Distribution by Capability Category}
\appcontentssub{app:tool_chaining}{B.2}{Emergent Tool-Chaining Behaviour}
\appcontentssub{app:tool_correctness}{B.3}{Tool Usage and Answer Correctness}
\appcontentssub{app:acc_tasktype}{B.4}{Accuracy by Task Type}
\appcontentssub{app:warn_provenance}{B.5}{Provenance of the diagnostic set}
\vspace{0.35em}
\appcontentsitem{sec:mining_procedure}{\textbf{C}}{Mining Procedure for Atomic Capabilities}
\appcontentssub{app:mining_pipeline}{C.1}{Step-by-Step Discovery Pipeline}
\vspace{0.35em}
\appcontentsitem{app:system_prompt}{\textbf{D}}{Agent System Prompt}
\endgroup
\vspace{0.8em}

\makeatletter
\let\app@origsection\section
\let\app@origsubsection\subsection
\renewcommand\section{\@startsection{section}{1}{\z@}%
  {-3.25ex \@plus -1ex \@minus -.2ex}%
  {1.5ex \@plus .2ex}%
  {\normalfont\large\bfseries}}
\renewcommand\subsection{\@startsection{subsection}{2}{\z@}%
  {-3.25ex \@plus -1ex \@minus -.2ex}%
  {1.5ex \@plus .2ex}%
  {\normalfont\normalsize\bfseries}}
\makeatother

\section{Capability Taxonomy}
\label{app:capability_taxonomy}

\begin{table*}[p]
    \centering
    \caption{Comprehensive taxonomy of 22 mined atomic capabilities for video understanding, grouped into five macro categories. The \textbf{Repr. Tool} column gives one representative tool; \textit{ReAct} denotes capabilities implemented through multi-step orchestration. Full tool sets and observed reasoning chains are given in Table~\ref{tab:mapping}.}
    \label{tab:capability_taxonomy}
    \footnotesize
    \renewcommand{\arraystretch}{1.05}
    \setlength{\tabcolsep}{4pt}
    \newcommand{\catrow}[2]{%
        \arrayrulecolor{black}\addlinespace[0.5pt]%
        \rowcolor{#1!18}\multicolumn{3}{@{}l}{\textcolor{#1!75!black}{\textbf{#2}}}\\[0.3pt]}
    \begin{tabularx}{\textwidth}{@{} >{\raggedright\arraybackslash}p{2.9cm} >{\raggedright\arraybackslash}p{3.2cm} >{\raggedright\arraybackslash}X @{}}
    \toprule
    \textbf{Atomic Capability} & \textbf{Representative Tool} & \textbf{Description} \\
    \midrule

    \catrow{taskblue}{Visual Perception}
    Visual Content Understanding & \texttt{inspect\_image} & Capturing the overall scene content, salient entities, and holistic semantic context of a frame or video segment. \\
    Object Perception & \texttt{detect} & Identifying object existence, spatial locations, and quantities. \\
    OCR / Text Perception & \texttt{ocr\_frame\_text} & Extracting and comprehending text embedded within visual frames. \\
    Action \& Event Understanding & \texttt{classify\_action} & Recognizing ongoing human/object actions and macro-events. \\
    Attribute \& State Analysis & \texttt{attribute\_recognition} & Analyzing fine-grained visual properties (e.g., color, texture) and scene classifications. \\
    Human-Centric Perception & \texttt{face\_detect} & Recognizing human biometrics including identity, pose, gaze, and facial expressions. \\
    \midrule

    \catrow{metricgreen}{Temporal Relation}
    Temporal Localization / Grounding & \texttt{locate\_event\_by\_text} & Identifying specific temporal occurrences, including query-based grounding and segment boundary detection using text embedding matching. \\
    Temporal Ordering / Sequencing & \texttt{caption\_video\_scenes} & Understanding chronological order and logical causal chains of events. \\
    State Change \& Transition & \texttt{scene\_change\_detect} & Detecting state transitions (e.g., object state transformations) and video shot boundaries. \\
    Temporal Continuity & \texttt{object\_track} & Tracking object persistence and estimating continuous action durations. \\
    Temporal Recognition & \texttt{classify\_action} & Defining and recognizing ongoing continuous actions or events over a time span. \\
    \midrule

    \catrow{spatorange}{Spatial Relation}
    Spatial Positioning & \texttt{locate} & Determining precise coordinates, bounding boxes, or existence of objects in 2D/3D space. \\
    Spatial Relationship & \texttt{spatial\_relation} & Understanding relative object arrangements (e.g., inside/outside), orientations, and occlusion states. \\
    Tracking \& Trajectory & \texttt{object\_track} & Capturing spatial movement paths and continuous displacement patterns of targets. \\
    Spatial Reasoning \& State & \texttt{estimate\_depth} & Inferring object states or making logical judgments based on complex spatial layouts and dynamic changes. \\
    \midrule

    \catrow{avpurple}{Audio-Visual}
    Speech Transcription (ASR) & \texttt{transcribe\_audio} & Transcribing spoken language content (Speech-to-Text). \\
    Audio-Visual Cross-modal & \texttt{detect\_active\_speaker} & Aligning visual and auditory signals temporally and semantically (e.g., active speaker detection); cross-modal fusion is handled as a ReAct workflow over these tools. \\
    Audio Event Detection & \texttt{detect\_audio\_events} & Identifying non-speech environmental sounds and acoustic events. \\
    \midrule

    \catrow{limitred}{High-Level Reasoning}
    Mathematical \& Quantitative & \textit{ReAct} & Performing arithmetic operations and numerical logic based on visual elements. \\
    Universal Counting & \textit{ReAct} & Quantifying discrete entities, actions, or temporal occurrences globally. \\
    Sequential \& Ordinal Reasoning & \textit{ReAct} & Deducing chronological sequences and causal implications among complex events. \\
    Abstract Logical Inference & \textit{ReAct} & Inferring macroscopic scene semantics or intentions based on implicit visual and spatial cues. \\

    \bottomrule
    \end{tabularx}
    \end{table*}

\subsection{Detailed Tool Mapping}
\label{app:tool_mapping}

\begin{table}[htbp]
\centering
\small
\renewcommand{\arraystretch}{1.2}
\setlength{\tabcolsep}{4pt}
\resizebox{\textwidth}{!}{%
\begin{tabular}{|c|l|p{12cm}|}
\hline
\multicolumn{3}{|c|}{\textbf{Visual Perception}} \\
\hline
\textbf{1} & Visual Content Understanding & \texttt{inspect\_image}, \texttt{query\_segment\_detail}, \texttt{caption\_video\_scenes} \\
\hline
\textbf{2} & Object Perception & \texttt{detect}, \texttt{locate}, \texttt{scene\_scan}, \texttt{count}, \texttt{instance\_segment} \\
\hline
\textbf{3} & OCR / Text Perception & \texttt{ocr\_frame\_text}, \texttt{ocr\_region\_read}, \texttt{ocr\_table\_extract}, \texttt{ocr\_video\_text\_timeline}, \texttt{ocr\_layout\_analysis} \\
\hline
\textbf{4} & Action \& Event Understanding & \texttt{classify\_action}, \texttt{classify\_action\_multi\_segments}, \texttt{recognize\_interactions} \\
\hline
\textbf{5} & Attribute \& State Analysis & \texttt{attribute\_recognition}, \texttt{prompt\_segment}, \texttt{instance\_segment} \\
\hline
\textbf{6} & Human-Centric Perception & \texttt{face\_detect}, \texttt{face\_cluster\_video}, \texttt{face\_match}, \texttt{analyze\_face\_emotion}, \texttt{gaze\_estimate}, \texttt{person\_identification} \\
\hline
\multicolumn{3}{|c|}{\textbf{Temporal Relation}} \\
\hline
\textbf{7} & Temporal Localization / Grounding & \texttt{locate\_event\_by\_text}, \texttt{locate\_event\_by\_scene\_caption}, \texttt{moment\_retrieve} \\
\hline
\textbf{8} & Temporal Ordering / Sequencing & \texttt{caption\_video\_scenes}, \texttt{classify\_action\_multi\_segments} \\
\hline
\textbf{9} & State Change \& Transition & \texttt{diff\_two\_frames}, \texttt{diff\_video\_timestamps}, \texttt{scene\_change\_detect} \\
\hline
\textbf{10} & Temporal Continuity & \texttt{object\_track}, \texttt{motion\_analyze}, \texttt{face\_cluster\_video} \\
\hline
\textbf{11} & Temporal Recognition & \texttt{classify\_action}, \texttt{ocr\_video\_text\_timeline}, \texttt{detect\_audio\_event\_timeline} \\
\hline
\multicolumn{3}{|c|}{\textbf{Spatial Relation}} \\
\hline
\textbf{12} & Spatial Positioning & \texttt{locate}, \texttt{spatial\_relation}, \texttt{calc\_bbox\_relation}, \texttt{calc\_point\_in\_region} \\
\hline
\textbf{13} & Spatial Relationship & \texttt{spatial\_relation}, \texttt{spatial\_analyze}, \texttt{calc\_multi\_spatial}, \texttt{calc\_bbox\_relation} \\
\hline
\textbf{14} & Tracking \& Trajectory & \texttt{object\_track}, \texttt{motion\_analyze}, \texttt{speed\_estimate}, \texttt{analyze\_flow\_segment} \\
\hline
\textbf{15} & Spatial Reasoning \& State & \texttt{estimate\_depth}, \texttt{compare\_object\_depth}, \texttt{get\_depth\_at\_points}, \texttt{spatial\_analyze} \\
\hline
\multicolumn{3}{|c|}{\textbf{Audio-Visual}} \\
\hline
\textbf{16} & Speech Transcription (ASR) & \texttt{transcribe\_audio}, \texttt{speaker\_diarize}, \texttt{get\_video\_subtitle} \\
\hline
\textbf{17} & Audio-Visual Cross-modal & \texttt{detect\_active\_speaker}, \texttt{analyze\_multimodal\_emotion}, \texttt{segment\_deep\_query}; \textit{Agent ReAct}: \texttt{get\_video\_subtitle} $\rightarrow$ \texttt{locate\_event\_by\_text} $\rightarrow$ \texttt{query\_segment\_detail} $\rightarrow$ LLM fusion \\
\hline
\textbf{18} & Audio Event Detection & \texttt{detect\_audio\_events}, \texttt{detect\_audio\_event\_timeline}, \texttt{identify\_music}, \texttt{compare\_audio\_fingerprint} \\
\hline
\multicolumn{3}{|c|}{\textbf{High-Level Reasoning}} \\
\hline
\textbf{19} & Mathematical \& Quantitative & \textit{ReAct Workflows}: such as \texttt{extract\_frame} $\rightarrow$ \texttt{ocr\_frame\_text} $\rightarrow$ \texttt{inspect\_image} $\rightarrow$ LLM arithmetic \\
\hline
\textbf{20} & Universal Counting & \textit{ReAct Workflows}: such as \texttt{generate\_thumbnail\_grid} $\rightarrow$ \texttt{inspect\_image} $\rightarrow$ \texttt{extract\_frame} $\rightarrow$ LLM aggregation \\
\hline
\textbf{21} & Sequential \& Ordinal Reasoning & \textit{ReAct Workflows}: such as \texttt{caption\_video\_scenes} $\rightarrow$ \texttt{query\_segment\_detail} $\rightarrow$ LLM temporal ordering \\
\hline
\textbf{22} & Abstract Logical Inference & \textit{ReAct Workflows}: such as \texttt{extract\_frame} $\rightarrow$ \texttt{concat\_images} $\rightarrow$ \texttt{inspect\_image} $\rightarrow$ LLM inference \\
\hline
\end{tabular}}
\caption{Mapping from the 22 canonical capabilities in Table~\ref{tab:capability_taxonomy} to concrete tools and observed multi-step chains. \textit{Agent ReAct} denotes multi-step orchestration; per-tool backends and descriptions are listed in Table~\ref{tab:tool_specs}.}
\label{tab:mapping}

\end{table}

\subsection{Backend Technologies}
\label{app:backend_technologies}

\begin{table}[htbp]
\centering
\renewcommand{\arraystretch}{1.2}
\begin{tabular}{|l|p{10cm}|}
\hline
\textbf{Backend} & \textbf{Description} \\
\hline
grounding-dino & Grounding-DINO~\cite{groundingdino:liu2024grounding} / DINO-X~\cite{dinox:ren2024dinox} for open-set detection, segmentation, and spatial analysis; DINO-X is used as the stronger grounding backend when available \\
\hline
vlm & Vision Language Models (GPT-4V, Gemini, etc.) for visual QA \\
\hline
whisperx & WhisperX~\cite{whisperx:bain2023whisperx} for word-level speech transcription \\
\hline
pyannote & Pyannote~\cite{pyannote:bredin2020neural} for speaker diarization \\
\hline
panns & Pretrained Audio Neural Networks for audio event detection \\
\hline
shazamio & Shazam-like music identification via audio fingerprinting \\
\hline
insightface & InsightFace for face detection, recognition, and emotion analysis \\
\hline
transformers & HuggingFace Transformers for action, emotion, depth models \\
\hline
sam & Segment Anything Model~\cite{sam:kirillov2023segment} for image segmentation \\
\hline
sam2 & SAM2~\cite{sam2:ravi2025sam2} for video object segmentation and tracking \\
\hline
clip & CLIP~\cite{clip:radford2021learning} for text-video moment retrieval \\
\hline
ffmpeg & FFmpeg for audio/video extraction and processing \\
\hline
script & Custom Python scripts for various utilities \\
\hline
yt-dlp & yt-dlp for YouTube video download and caching \\
\hline
\end{tabular}
\caption{Backend Technology Stack}
\label{tab:backend_technologies}
\end{table}

\newpage

\subsection{Tool Specifications}
\label{app:tool_specs}

\begingroup
\footnotesize
\renewcommand{\arraystretch}{1.18}
\setlength{\tabcolsep}{4pt}
\setlength{\LTcapwidth}{\textwidth}
\renewcommand{\_}{\char`\_\allowbreak}
\begin{longtable}{| >{\raggedright\arraybackslash}p{3.5cm} | >{\raggedright\arraybackslash}p{2.7cm} | >{\raggedright\arraybackslash}p{8.6cm} |}
\caption{Tool specifications: backend and function (70 tools).}
\label{tab:tool_specs} \\
\hline
\textbf{Tool} & \textbf{Backend} & \textbf{Description} \\
\hline
\endfirsthead
\caption[]{Tool specifications (continued).} \\
\hline
\textbf{Tool} & \textbf{Backend} & \textbf{Description} \\
\hline
\endhead
\hline
\endfoot
\hline
\endlastfoot
\multicolumn{3}{|c|}{\textbf{VLM Query}} \\
\hline
\texttt{inspect\_image} & vlm & Visual QA on a frame, grid, or composite image. \\
\hline
\texttt{query\_segment\_detail} & vlm & VLM QA over a time interval using sampled frames and optional SRT. \\
\hline
\texttt{caption\_video\_scenes} & vlm & Timestamped captions for detected scenes. \\
\hline
\multicolumn{3}{|c|}{\textbf{Grounding-DINO Visual Detection}} \\
\hline
\texttt{detect} & grounding-dino & Open-set detection of named categories with boxes and scores. \\
\hline
\texttt{locate} & grounding-dino & Ground a natural-language phrase to matching boxes. \\
\hline
\texttt{scene\_scan} & grounding-dino & Scan common categories to build a scene inventory. \\
\hline
\texttt{count} & grounding-dino & Count instances of a named category. \\
\hline
\texttt{instance\_segment} & grounding-dino & Detect objects and report relative instance size. \\
\hline
\texttt{spatial\_relation} & grounding-dino & Detect two objects and compute their geometric relation. \\
\hline
\multicolumn{3}{|c|}{\textbf{OCR Text Extraction}} \\
\hline
\texttt{ocr\_frame\_text} & PaddleOCR & Extract on-screen text with boxes and confidence. \\
\hline
\texttt{ocr\_region\_read} & PaddleOCR & Read text from a specified image region. \\
\hline
\texttt{ocr\_table\_extract} & PaddleOCR & Parse tables and scoreboards into structured fields. \\
\hline
\texttt{ocr\_video\_text\_timeline} & PaddleOCR & Track when overlay text appears or changes. \\
\hline
\texttt{ocr\_layout\_analysis} & PaddleOCR & Segment document or slide regions and infer reading order. \\
\hline
\multicolumn{3}{|c|}{\textbf{Action Recognition}} \\
\hline
\texttt{classify\_action} & transformers & Classify the dominant activity in a clip. \\
\hline
\texttt{classify\_action\_multi\_segments} & transformers & Classify actions across multiple time windows. \\
\hline
\texttt{recognize\_interactions} & grounding-dino & Detect entities and action-relation triples. \\
\hline
\multicolumn{3}{|c|}{\textbf{Face \& Emotion Analysis}} \\
\hline
\texttt{face\_detect} & insightface & Detect faces with boxes and scores. \\
\hline
\texttt{face\_cluster\_video} & insightface & Cluster faces across frames to identify unique people. \\
\hline
\texttt{face\_match} & insightface & Verify identity correspondence between two face images. \\
\hline
\texttt{analyze\_face\_emotion} & insightface & Classify facial expressions into emotion categories. \\
\hline
\texttt{analyze\_speech\_emotion} & transformers & Classify vocal emotion in an audio or video span. \\
\hline
\texttt{analyze\_multimodal\_emotion} & insightface & Fuse facial and speech emotion at a timestamp. \\
\hline
\texttt{gaze\_estimate} & insightface & Estimate gaze direction and coarse target. \\
\hline
\texttt{detect\_active\_speaker} & insightface & Identify the active speaker from face and audio cues. \\
\hline
\multicolumn{3}{|c|}{\textbf{Visual Attribute Recognition}} \\
\hline
\texttt{attribute\_recognition} & vlm & Query visual attributes globally or within boxes. \\
\hline
\texttt{person\_identification} & vlm & Match people across frames using visual appearance. \\
\hline
\multicolumn{3}{|c|}{\textbf{Temporal Analysis}} \\
\hline
\texttt{locate\_event\_by\_text} & script & Rank subtitle spans matching a text query. \\
\hline
\texttt{moment\_retrieve} & clip & Retrieve visually matching moments. \\
\hline
\texttt{scene\_change\_detect} & script & Detect shot boundaries from frame differences. \\
\hline
\multicolumn{3}{|c|}{\textbf{Spatial Analysis}} \\
\hline
\texttt{calc\_bbox\_relation} & script & Compute direction, overlap, containment, and distance between boxes. \\
\hline
\texttt{calc\_point\_in\_region} & script & Test whether a point or box center lies inside a region. \\
\hline
\texttt{calc\_multi\_spatial} & script & Compute spatial relations for multiple objects. \\
\hline
\texttt{spatial\_crop} & script & Crop a specified image region for inspection. \\
\hline
\texttt{spatial\_analyze} & grounding-dino & Detect categories and summarize pairwise layout. \\
\hline
\multicolumn{3}{|c|}{\textbf{Instance Segmentation \& Tracking}} \\
\hline
\texttt{auto\_segment} & sam & Segment distinct image regions without prompts. \\
\hline
\texttt{prompt\_segment} & sam & Generate a mask from point or box prompts. \\
\hline
\texttt{tracked\_object\_count} & sam2 & Count tracked instances over a short clip. \\
\hline
\texttt{object\_track} & sam2 & Track a prompted object across frames. \\
\hline
\multicolumn{3}{|c|}{\textbf{Audio Processing}} \\
\hline
\texttt{transcribe\_audio} & whisperx & Transcribe speech with word timestamps. \\
\hline
\texttt{speaker\_diarize} & pyannote & Assign speaker labels to audio turns. \\
\hline
\texttt{detect\_audio\_events} & panns & Classify non-speech audio events. \\
\hline
\texttt{detect\_audio\_event\_timeline} & panns & Produce a timestamped audio-event timeline. \\
\hline
\texttt{identify\_music} & shazamio & Identify music from an audio excerpt. \\
\hline
\texttt{compare\_audio\_fingerprint} & shazamio & Compare two audio fingerprints for matching content. \\
\hline
\multicolumn{3}{|c|}{\textbf{Video Basic Operations}} \\
\hline
\texttt{get\_video\_duration} & script & Return video duration, resolution, and stream metadata. \\
\hline
\texttt{get\_videos\_duration} & script & Return durations for multiple videos. \\
\hline
\texttt{get\_directory\_videos\_duration} & script & Scan a directory for video durations. \\
\hline
\texttt{extract\_frame} & script & Extract frames at specified timestamps. \\
\hline
\texttt{generate\_thumbnail\_grid} & script & Build a timestamped thumbnail grid. \\
\hline
\texttt{extract\_audio} & ffmpeg & Extract or resample the audio track. \\
\hline
\texttt{extract\_video\_segment} & ffmpeg & Extract a temporal video segment. \\
\hline
\texttt{get\_video\_subtitle} & script & Retrieve SRT text for a video or time range. \\
\hline
\texttt{get\_multiple\_videos\_subtitle} & script & Retrieve subtitles for multiple videos. \\
\hline
\texttt{concat\_images} & script & Combine frames into a comparison grid. \\
\hline
\texttt{check\_subtitle\_exists} & script & Check whether subtitles are available. \\
\hline
\texttt{get\_video\_from\_youtube\_id} & yt-dlp & Resolve a YouTube ID to a local video path. (This is only for MINERVA testing.) \\
\hline
\texttt{check\_youtube\_cache} & yt-dlp & Check the local YouTube cache. \\
\hline
\multicolumn{3}{|c|}{\textbf{Motion \& Depth Analysis}} \\
\hline
\texttt{diff\_two\_frames} & script & Compute pixel differences between two frames. \\
\hline
\texttt{diff\_video\_timestamps} & script & Compare frames at two video timestamps. \\
\hline
\texttt{motion\_analyze} & sam2 & Summarize the motion of a tracked object. \\
\hline
\texttt{speed\_estimate} & script & Estimate speed from an object trajectory. \\
\hline
\texttt{compute\_dense\_flow} & script & Compute dense optical flow between frames. \\
\hline
\texttt{analyze\_flow\_segment} & script & Summarize optical flow over a video segment. \\
\hline
\texttt{estimate\_depth} & transformers & Estimate a monocular depth map. \\
\hline
\texttt{compare\_object\_depth} & transformers & Compare the depth of two detected objects. \\
\hline
\texttt{get\_depth\_at\_points} & transformers & Sample depth at specified image points. \\
\hline
\multicolumn{3}{|c|}{\textbf{Camera Motion Analysis}} \\
\hline
\texttt{analyze\_camera\_motion} & script & Classify camera motion over a video clip. \\
\hline
\texttt{detect\_camera\_motion\_type} & script & Classify camera motion between two timestamps. \\
\hline
\end{longtable}
\endgroup

\newpage

\section{Tool Usage Distribution Analysis}
\label{app:tool_usage}

\subsection{Usage Distribution by Capability Category}
\label{app:usage_by_category}

\begin{figure}[H]
  \centering
  \includegraphics[width=\textwidth]{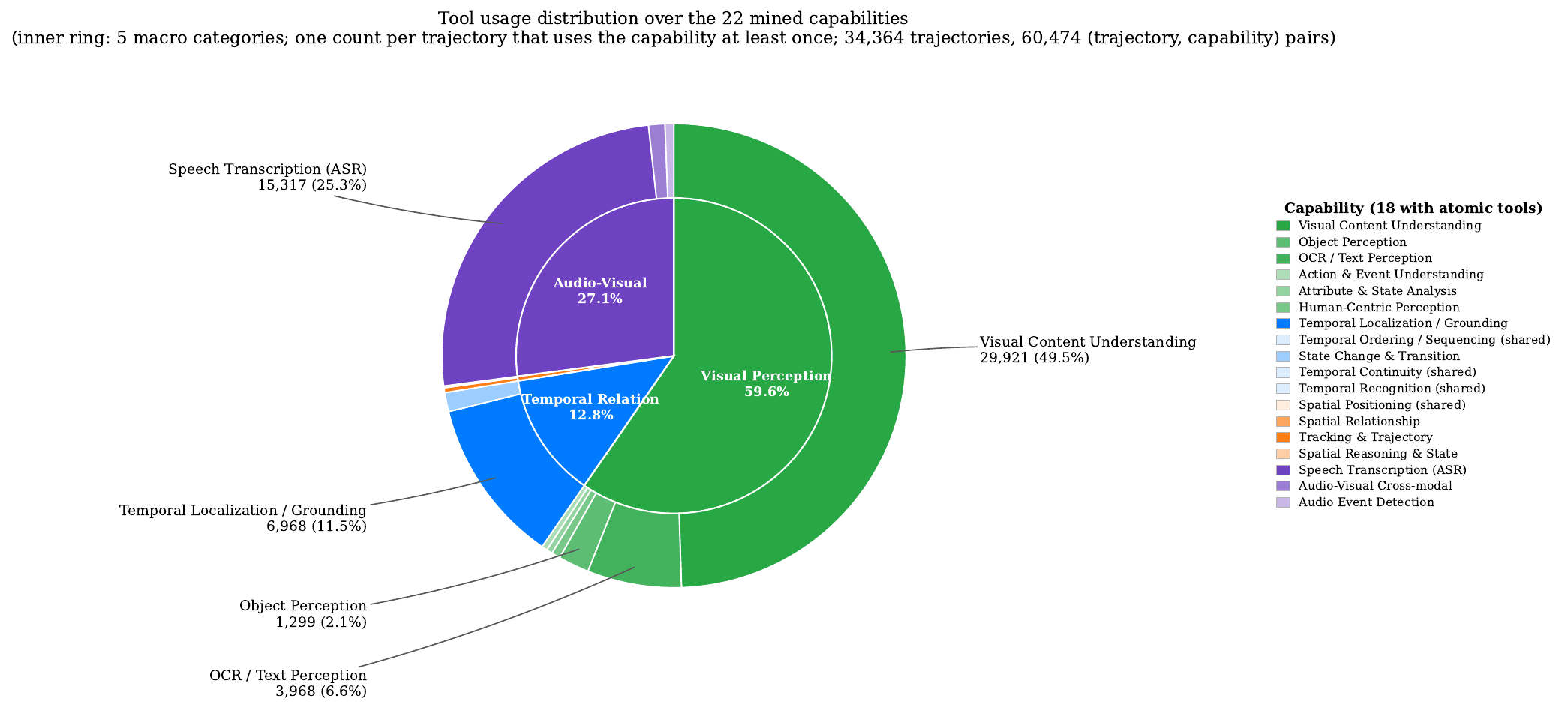}
  \caption{Tool usage distribution over the 22 mined capabilities, counted once per trajectory
  that exercises the capability. The inner ring aggregates the five macro categories; the outer
  ring shows individual capabilities, shaded within their macro colour. Percentages are shares
  of the 60,474 (trajectory, capability) pairs. The legend covers the 18 capabilities that own at
  least one atomic tool in Table~\ref{tab:mapping}; capabilities 19--22 are realized only as ReAct
  chains and are therefore not measurable in a tool-level count. A pale swatch marks a capability
  with no mass (\textsf{shared}): its tools are also listed under another capability and are
  credited there, so it may be exercised in practice but not counted here.}
  \label{fig:capability_pie}
  \end{figure}

Figure~\ref{fig:capability_pie} reports how often each of the 22 mined capabilities is
actually exercised. The counting unit here is deliberately a \emph{trajectory} rather than a
tool call: every concrete tool invoked in a trajectory is mapped back to its capability
through the correspondence of Table~\ref{tab:mapping}, and a capability is credited once if
the trajectory touched \emph{any} of its tools at least once, no matter how often. This
removes the bias of chatty inner loops---a single sample may call \texttt{inspect\_image}
dozens of times---so the figure measures capability \emph{coverage} instead of total call volume.
Generic operators that carry no capability semantics (e.g.\ \texttt{get\_video\_duration},
\texttt{extract\_frame}). Over the full transcript corpus of
partially collected 34,364 trajectories, 33,544 (97.6\%) exercise at least one capability of
Table~\ref{tab:mapping}, giving 60,474 (trajectory, capability) pairs.

The distribution is strongly head-heavy: Visual Content Understanding accounts for 49.5\% of
all pairs (present in 87.1\% of trajectories) and Speech Transcription for 25.3\% (44.6\% of
trajectories), followed by temporal localization (11.5\%) and OCR (6.6\%); the remaining ten
capabilities together contribute under 8\%.

Eight capabilities carry no mass in the figure, for three different reasons, and neither means the
capability is unexercised. (1) Capabilities 19--22 own no atomic tool at all: Table~\ref{tab:mapping}
realizes them purely as ReAct chains over other tools, so they lie outside what a tool-level count
can measure. (2) For capabilities 8, 10, 11 and 12, their tools are also listed under another capability and are
credited there, so it is exercised in practice but not counted here. (3) The trajectories for statistics is partially collected from all dates, so some tools may be omitted.

Moreover, a small share should not be read as low utility. Frequency here reflects how often a
capability is \emph{needed}, not how much it contributes when it is: a long-tail tool is
typically the only route to the evidence a particular question depends on, so its value is
whether those few queries become answerable at all, not the fraction of trajectories it
appears in. The ablation in Table~\ref{tab:ablation_study} makes this concrete---the visual
expert group (detection, OCR, segmentation and tracking) accounts for roughly a tenth of the
pairs in this figure, yet removing it costs $3.33$ accuracy points.

\subsection{Emergent Tool-Chaining Behaviour}
\label{app:tool_chaining}

\begin{figure}[H]
  \centering
  \includegraphics[width=0.86\textwidth]{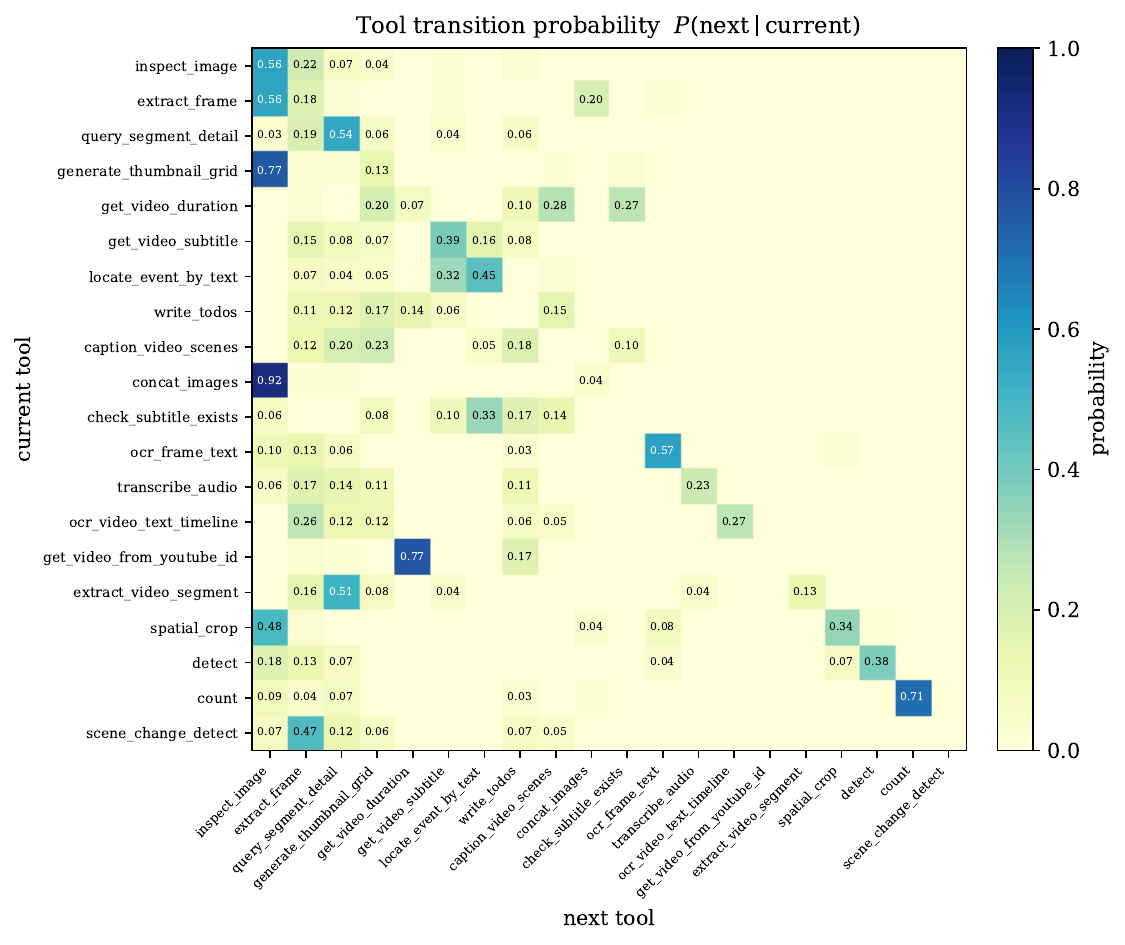}
  \caption{First-order tool transition probabilities $P(\mathrm{next}\mid\mathrm{current})$
  for the twenty most-frequent tools, i.e.\ the high-traffic backbone ($99.3\%$ of all calls);
  this is a view of the common path, not a ranking of tool importance.
  Rows sum to~1 over the full tool set; only each tool's six most likely successors are
  retained (together $92$--$99\%$ of its outgoing mass), and of those only the cells landing
  inside the top~20 are shown.
  Bright diagonal/off-diagonal cells expose the emergent perception loop and the subtitle$\leftrightarrow$grounding cycle.}
  \label{fig:tool_transition}
  \end{figure}

Beyond marginal call counts, the ordered traces reveal how the agent
\emph{composes} tools.
Figure~\ref{fig:tool_transition} shows the first-order Markov transition matrix $P(\mathrm{next}\mid\mathrm{current})$ over the twenty most-used tools.
Two macro-workflows emerge without being hard-coded:
(i)~a \textbf{frame-perception loop} in which \texttt{generate\_thumbnail\_grid}, \texttt{extract\_frame}, and \texttt{concat\_images} all funnel into \texttt{inspect\_image} ($P\!\ge\!0.55$, up to $0.92$ from \texttt{concat\_images}), which then largely self-loops or hands off to \texttt{query\_segment\_detail};
and (ii)~a \textbf{text/temporal-grounding chain} \texttt{get\_video\_duration}$\rightarrow$\texttt{caption\_video\_scenes}/\texttt{check\_subtitle\_exists} ($0.28$/$0.27$) followed by a subtitle$\leftrightarrow$grounding cycle between \texttt{get\_video\_subtitle} and \texttt{locate\_event\_by\_text} ($0.16$ and $0.32$ in the two directions), where both tools also self-loop strongly ($0.39$ and $0.45$) as the agent walks successive subtitle windows.
These two clusters correspond to the visual and language/temporal evidence-gathering strategies.

This is a traffic-weighted view: the twenty tools shown carry $99.3\%$ of the $173{,}846$ calls, so
the matrix describes the path the agent takes on typical questions---not a ranking of which tools
matter. A specialist tool is called only when a question happens to need it, so it can never form a
high-probability edge however useful it is. Their value shows up in the ablation study
ablation of Table~\ref{tab:ablation_study}.

\subsection{Tool Usage and Answer Correctness}
\label{app:tool_correctness}

\begin{figure}[H]
\centering
\includegraphics[width=0.78\textwidth]{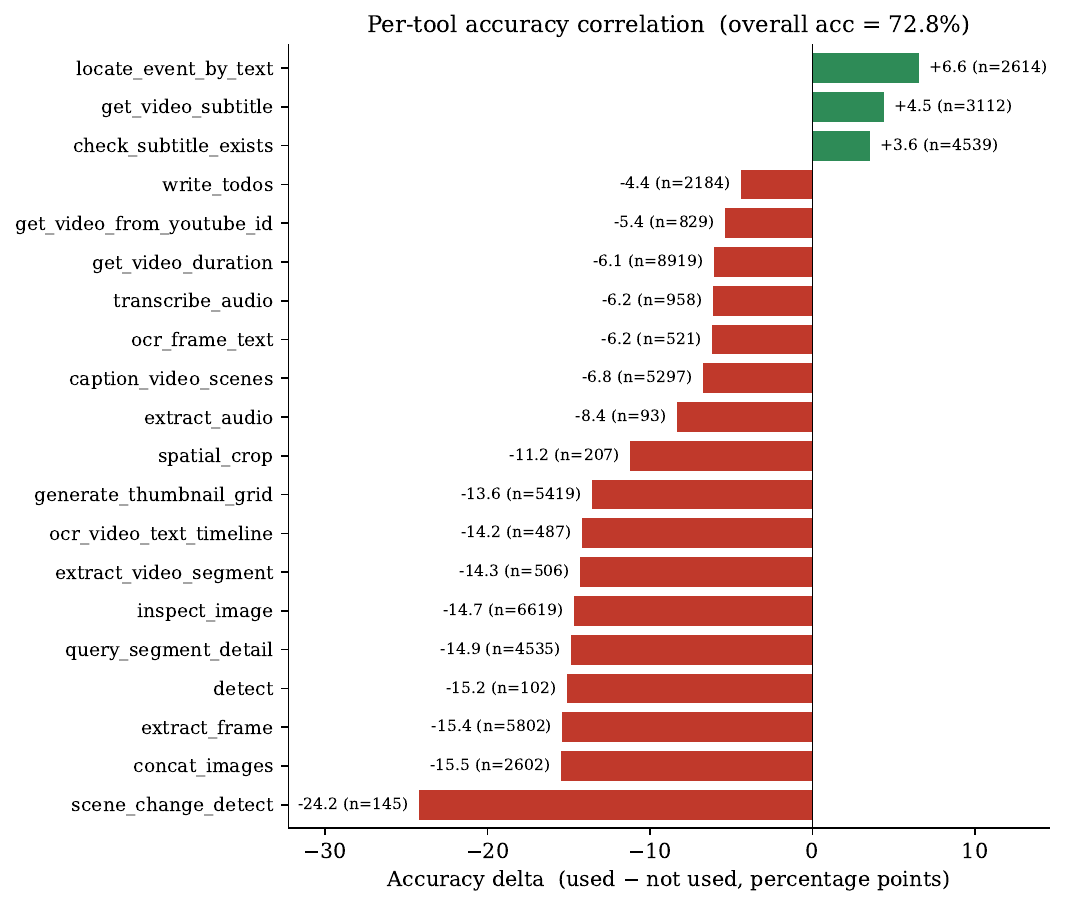}
\caption{Per-tool accuracy correlation.
Green = sessions using the tool score \emph{above} the no-use baseline; red = below.
Bars are annotated with $\Delta$ (percentage points) and $n_{\text{used}}$.
Overall accuracy $=72.8\%$.}
\label{fig:tool_delta}
\end{figure}

Figure~\ref{fig:tool_delta} contrasts, for each tool, the accuracy of sessions that \emph{used} it against those that did not ($\Delta = \mathrm{acc_{used}} - \mathrm{acc_{not}}$; only tools with $n_{\text{used}}\!\ge\!20$ and support~$\ge\!1\%$ are shown).
These observational correlations are not causal effects: harder questions tend to trigger \emph{more} tool calls, so heavy visual tools (e.g.\ \texttt{inspect\_image}, \texttt{extract\_frame}, \texttt{query\_segment\_detail}, all $\Delta\!\approx\!-15$pp) appear negatively associated mainly because they are invoked on the hardest items.
The only positively associated tools are the subtitle/temporal-grounding family (\texttt{locate\_event\_by\_text} $+6.6$pp, \texttt{get\_video\_subtitle} $+4.5$pp, \texttt{check\_subtitle\_exists} $+3.6$pp), reflecting that questions answerable from transcripts are both easier and text-tool-driven.

At the \emph{pair} level, the same signal sharpens: co-using the subtitle and temporal-grounding tools yields the largest positive lift, whereas pairs anchored on \texttt{scene\_change\_detect} or \texttt{ocr\_video\_text\_timeline} mark the hardest slices.
Here $\Delta$ is measured against the \emph{overall} baseline (not a leave-one-pair-out control).

\begin{table}[htbp]
\centering
\renewcommand{\arraystretch}{1.25}
\small
\begin{tabular}{|l|r|r|r|}
\hline
\textbf{Tool Pair} & \textbf{$n$} & \textbf{Acc.} & \textbf{$\Delta$ vs.\ overall} \\
\hline
\multicolumn{4}{|c|}{\cellcolor{metricgreen!15}\textcolor{metricgreen}{\textbf{Most positively associated}}} \\
\hline
\texttt{get\_video\_subtitle} + \texttt{locate\_event\_by\_text} & 1,897 & 79.1\% & \textcolor{metricgreen}{$+6.2$pp} \\
\hline
\texttt{check\_subtitle\_exists} + \texttt{locate\_event\_by\_text} & 2,322 & 78.2\% & \textcolor{metricgreen}{$+5.4$pp} \\
\hline
\texttt{get\_video\_duration} + \texttt{locate\_event\_by\_text} & 2,500 & 77.4\% & \textcolor{metricgreen}{$+4.6$pp} \\
\hline
\texttt{check\_subtitle\_exists} + \texttt{get\_video\_subtitle} & 2,632 & 76.7\% & \textcolor{metricgreen}{$+3.8$pp} \\
\hline
\multicolumn{4}{|c|}{\cellcolor{limitred!15}\textcolor{limitred}{\textbf{Most negatively associated}}} \\
\hline
\texttt{locate\_event\_by\_text} + \texttt{ocr\_video\_text\_timeline} & 115 & 47.8\% & \textcolor{limitred}{$-25.0$pp} \\
\hline
\texttt{generate\_thumbnail\_grid} + \texttt{scene\_change\_detect} & 133 & 48.1\% & \textcolor{limitred}{$-24.7$pp} \\
\hline
\texttt{get\_video\_subtitle} + \texttt{ocr\_video\_text\_timeline} & 103 & 48.5\% & \textcolor{limitred}{$-24.3$pp} \\
\hline
\texttt{inspect\_image} + \texttt{scene\_change\_detect} & 144 & 48.6\% & \textcolor{limitred}{$-24.2$pp} \\
\hline
\end{tabular}
\caption{Tool-pair correctness correlation for sessions using \emph{both} tools.
$\Delta$ is relative to the overall $72.8\%$ baseline.}
\label{tab:pair_correctness}
\end{table}

\paragraph{Interpretation and confounds.}

The sign pattern is consistent: positive associations are concentrated among text-based tools (subtitle / temporal grounding), whereas heavy visual tools are negatively associated.
This is not evidence that ``looking at the video hurts.''
It is driven by two confounds that operate in opposite directions.

\textbf{(1) Visual tools are invoked on harder questions.}
The agent decides which tools to call, and it escalates to visual inspection precisely when a question resists an easy answer.
Figure~\ref{fig:tool_delta} makes this concrete: \texttt{inspect\_image}, \texttt{query\_segment\_detail}, \texttt{concat\_images}, and especially \texttt{scene\_change\_detect}---the most negative tool---are the ones reserved for items that resist a short, text-first pass.
The visual tools' baseline $\mathrm{acc_{not}}$ is therefore enriched for shorter sessions that terminate after only a few calls; this imbalance can inflate the reference accuracy and induce a negative association in $\Delta$.
By contrast, subtitle tools appear across ordinary sessions rather than only on that leftover easy tail, so their positive $\Delta$ is not an artefact of short-session dilution.

\textbf{(2) Very common tools yield unstable comparisons.}
For a tool used in almost every session (e.g.\ \texttt{get\_video\_duration}, support~$0.97$), the ``did-not-use'' comparison group is a tiny, atypical $3\%$ sliver, so its $\Delta$ is statistically fragile and should not be over-read.

\textbf{(3) Subtitle use remains associated after stratification.}
To separate subtitle availability from subtitle use, we restrict the analysis to the $4{,}539$ sessions that probed for subtitles and compare those that then \emph{read} the transcript with those that did not.
The lift remains in this matched stratum: \texttt{get\_video\_subtitle} $+4.8$pp ($76.7\%$ vs.\ $71.8\%$) and \texttt{locate\_event\_by\_text} $+7.3$pp ($78.2\%$ vs.\ $70.9\%$).
This residual is consistent with a modality difference: a transcript provides a high-fidelity text channel, whereas visual perception can be affected by sparse frame sampling and VLM captioning errors.

\textbf{Summary.}
The correlations largely measure \emph{which questions summon which tools}, not the marginal value of a tool.
Visual tools are negatively associated because they are the instruments of last resort on the hardest items; subtitle tools are positively associated both because they select for text-answerable questions and because, controlling for availability, text remains the cleanest evidence channel.

\subsection{Accuracy by Task Type}
\label{app:acc_tasktype}

\begin{figure}[H]
\centering
\includegraphics[width=0.82\textwidth]{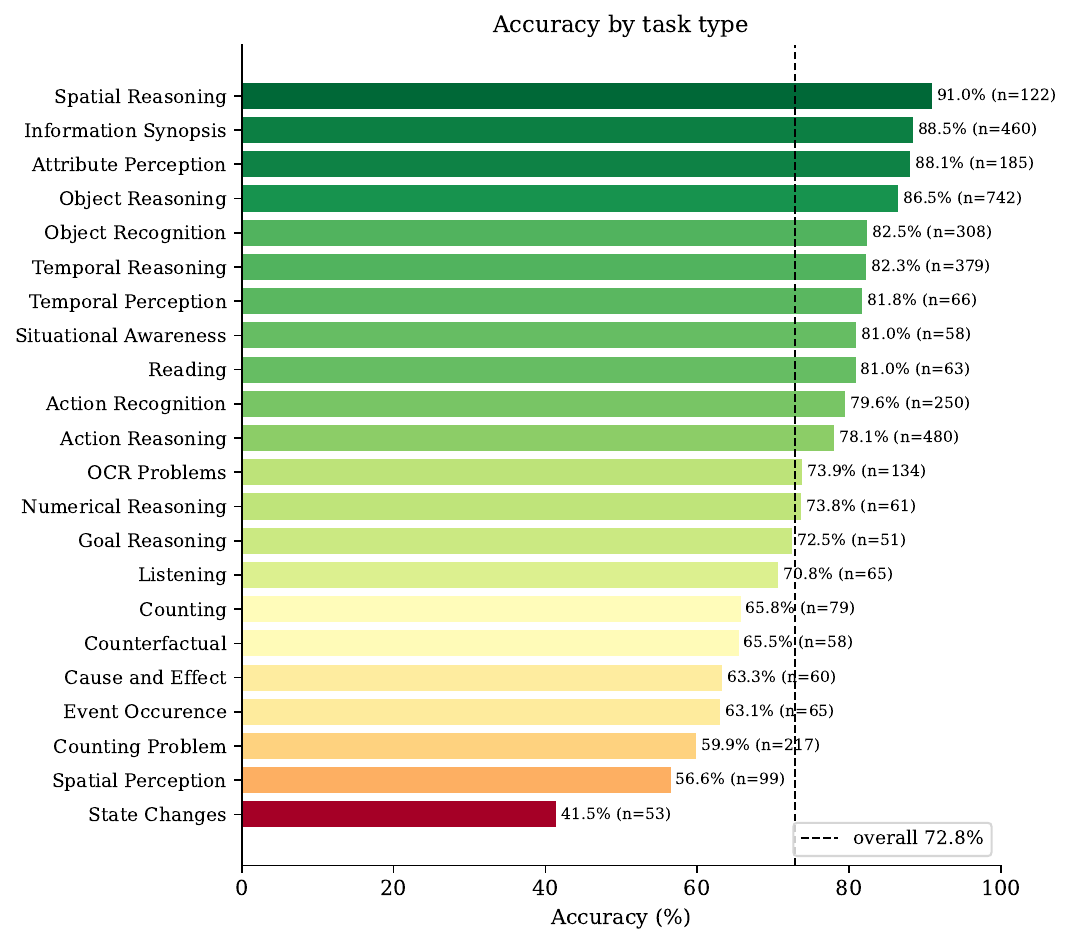}
\caption{Accuracy per task type (labelled sessions with $n\!\ge\!50$).
Dashed line = overall $72.8\%$.
Colour encodes accuracy; annotations give accuracy and sample size.}
\label{fig:acc_tasktype}
\end{figure}

Aggregating outcomes by question type (Figure~\ref{fig:acc_tasktype}) makes the tool$\leftrightarrow$difficulty link concrete: the strongest categories (\emph{Spatial Reasoning} $91\%$, \emph{Information Synopsis} $88\%$, \emph{Attribute Perception} $88\%$, \emph{Object Reasoning} $87\%$) are dominated by subtitle/temporal tools and use few calls per session, whereas the weakest (\emph{State Changes} $42\%$, \emph{Spatial Perception} $57\%$, \emph{Counting} $60\%$) rely on dense frame inspection and issue the most calls.

\subsection{Provenance of the diagnostic set}
\label{app:warn_provenance}
The 217 loop-prone questions used in the budget-control ablation
(Sec.~\ref{sec:abl_warning}) are the union of all cases on which an \emph{independent} agent
backbone (Gemini-3.1-Pro) exhausted its step budget without producing a prediction, collected
across VideoMME, VideoMME-v2, MINERVA, LVBench, and LongVideoBench. We reuse only the
\emph{inputs}; every run reported in the ablation uses the Kimi-K2.6 backbone, so the
selection is not circular with respect to the evaluated system. This set is therefore a
targeted stress test for the non-termination failure mode, not a random sample of the
benchmarks; benchmark-wide results appear in Table~\ref{tab:sota_comparison}.

\section{Mining Procedure for Atomic Capabilities}
\label{sec:mining_procedure}
For reproducibility, we represent each video-QA instance as the quadruple:
\begin{equation}
d = (Q, R, A, O)
\end{equation}
where $Q$ is the query for video $V$, $R$ is the expert reasoning trace, $A$ is the ground-truth answer, and $O$ is the set of candidate options.

Let $\mathcal{C} = \{c_1, c_2, \dots, c_K\}$ denote the global taxonomy of standardized \textit{atomic capabilities}; the capabilities required by instance $d$ form a subset:
\begin{equation}
\mathcal{C}_d \subseteq \mathcal{C}
\end{equation}

The pipeline below extracts and aggregates these subsets from $R$, then maps $\mathcal{C}_d$ to executable tools $T$ under budget $B$ (i.e., $A = \operatorname{Agent}(V, Q, T, B)$ as in Sec.~\ref{sec:problem_formulation}).

\subsection{Step-by-Step Discovery Pipeline}
\label{app:mining_pipeline}
This section specifies the implementation of four phases: (1) LLM prompting, (2) merging and filtering, (3) embedding and clustering, and (4) final taxonomy construction.

To illustrate the trace-level decomposition, consider the paraphrased sports query in Table~\ref{tab:example_qa}. A coarse label such as ``Sports Analysis'' hides several operations in the reasoning trace: reading the scoreboard, identifying teams or players from jersey attributes, recognizing basketball actions, and linking those actions to the observed score change. These raw descriptors are subsequently normalized, clustered, and mapped to the canonical capability names reported in Table~\ref{tab:capability_taxonomy}.

\begin{table}[t]
\caption{Paraphrased schematic of a MINERVA-style $(Q,R,A,O)$ record, used only to illustrate capability extraction. The query is shortened, options are collapsed, and the video key is omitted; quoted fragments of $R$ are representative rather than a reproduced benchmark item.}
\label{tab:example_qa}
\centering
\small
\begin{tabular}{lp{6.2cm}}
\toprule
\textbf{Field} & \textbf{Content} \\
\midrule
\textbf{Domain} & Sports (ball game) \\
\textbf{Question type} & Event occurrence \\
\midrule
\textbf{Question (Q)} & (paraphrased) Which on-court sequence caused one team's score to become $24$--$13$? \\
\midrule
\textbf{Options (O)} & Collapsed into two types: a sequence ending in a three-point shot vs.\ sequences ending in a dunk, fadeaway, or layup. \\
\midrule
\textbf{Quoted from $R$} & ``read the scoreboard and added 3''; temporal grounding of the score-change moment; visual identification of jersey numbers and a rebound-then-pass. \\
\midrule
\textbf{Mined capabilities} & OCR / Text Perception; Temporal Localization; Mathematical \& Quantitative. \\
\bottomrule
\end{tabular}
\end{table}

\begin{figure}[htbp]
\centering
\begin{lstlisting}[
    basicstyle=\scriptsize\ttfamily,   % 等宽小字体
    frame=single,                     % 细边框
    backgroundcolor=\color{gray!5},   % 淡淡的灰色背景
    breaklines=true                  % 自动折行
]
# Role
You are an expert AI Analyst specializing in long video Understanding. Your task is to reverse-engineer the cognitive process required to solve a specific video QA example.

# Task
Analyze the provided [Question], [Final Answer], [Reasoning], and [Distractors]. Determine the specific atomic capabilities required to answer the question correctly.
- An atomic capability is the smallest practically distinguishable component of a skill.

# Reference Taxonomy (Start Here)
Check if the required ability fits into these categories. Give a specific name and attribute it to a category. If NOT, you must define a new category.
1. Visual Perception
2. Temporal Dynamics
3. Spatial & Object Dynamics
4. Audio-Visual
5. High-Level Reasoning

# Protocol for Defining New Abilities (Extension Rules)
If the problem-solving requires a capability strictly outside the list above:
1. Create a precise name (e.g., Emotion Recognition).
2. Assign a Category.
3. Justify explicitly why existing tags were insufficient.

# JSON Schema Output Requirement
{
  "analysis_summary": {
    "primary_composite_skill": "String",
    "complexity_level": "Low | Medium | High | Very High",
    "composite_skill_description": "String"
  },
  "required_atomic_abilities": [
    {
      "ability_name": "String",
      "category": "String",
      "definition_justification": "String",
      "reasoning_evidence": "String (Quote text from Reasoning)",
      "necessity": "Required | Optional"
    }
  ]
}
\end{lstlisting}
\caption{The system prompt (\texttt{Thinking\_prompt}) used to guide the LLM in reverse-engineering human reasoning traces.}
\label{fig:thinking_prompt}
\end{figure}

\subsubsection{LLM prompting}
The first phase of our pipeline systematically extracts raw, fine-grained capability descriptors from human-annotated reasoning traces. Specifically, each task quadruple $d = (Q, R, A, O)$ is formatted as a structured textual input. Table~\ref{tab:example_qa} shows a paraphrased schematic of such a record: solving the question typically combines temporal grounding, visual OCR, and basic arithmetic.

To guide the LLM in systematically decomposing this cognitive trace, we design a dedicated system prompt, denoted as \texttt{Thinking\_prompt} (detailed in Fig.~\ref{fig:thinking_prompt}). To reasonably anchor the open-ended extraction space, we provide the LLM with five empirical prior-knowledge macro-categories, derived from existing video benchmarks and dataset characteristics:

\begin{enumerate}
    \item \textbf{Visual Perception}: Lower-level visual recognition tasks, such as object detection, attribute identification, and Optical Character Recognition (OCR).
    \item \textbf{Temporal Dynamics}: Time-related operations, including temporal grounding (identifying timestamps) and temporal event ordering.
    \item \textbf{Spatial \& Object Dynamics}: Spatial relation understanding and object counting within frames.
    \item \textbf{Audio-Visual}: Multi-sensory comprehension, such as speech-to-text transcription and cross-modal audio-visual alignment.
    \item \textbf{High-Level Reasoning}: Logical deductions, basic arithmetic calculations, and counterfactual scenario planning.
\end{enumerate}

During inference, the LLM reads the input instance and reverse-engineers the discrete steps human experts took to solve the task. The model is forced to map its findings into a strict JSON schema containing (1) a high-level \texttt{analysis\_summary} of the composite skill and (2) a list of \texttt{required\_atomic\_abilities}.

Crucially, to prevent the LLM from hallucinating or generating unrealistic capability names, the schema enforces a \texttt{reasoning\_evidence} field for every extracted capability. The LLM must populate this field by quoting direct text segments from the human reasoning trace $R$. For instance, in the schematic of Table~\ref{tab:example_qa}, the phrase ``read the scoreboard and added 3'' maps to \textit{Arithmetic} (under High-Level Reasoning) with the quoted fragment as validation. This grounded mechanism ensures that every extracted raw capability is strictly backed by the factual steps of human experts.

\subsubsection{Merging \& Filtering}
The raw capabilities extracted directly from the LLM inevitably contain syntactic redundancies and exhibit a long-tail distribution. To address this, we implement a two-stage filtering pipeline: syntactic normalization followed by frequency-based semantic merging.

First, we resolve surface-level formatting variations by applying a deterministic normalization function. This step converts all raw ability names to lowercase, replaces hyphens and underscores with spaces, expands parentheses, and applies a domain-specific vocabulary mapping to unify common variations (e.g., mapping both ``\textit{fine-grained}'' and ``\textit{fine grained}'' to the unified term ``\textit{finegrained}''). In our dataset, this lightweight step successfully identified and merged eight groups of syntactic variants, which primarily suffered from inconsistent capitalization and hyphenation styles (such as \textit{Fine-Grained} versus \textit{Fine-grained}).
Second, to filter the sparse noise inherent in the long-tail distribution, we apply a frequency-based merging strategy with a minimum occurrence threshold of 5, identifying 458 low-frequency terms. To evaluate semantic equivalence, we encode all terms into dense 384-dimensional vectors using a locally deployed \texttt{all-MiniLM-L6-v2} Sentence Transformer~\cite{minilm:wang2020minilm,sbert:reimers2019sentencebert}. We compute the cosine similarity between the embeddings of these low-frequency terms and all high-frequency terms, merging each rare capability into the target with the highest similarity. For instance, the low-frequency term \textit{Person Re-Identification} (frequency of 2) is mapped to \textit{Person Re-identification} (similarity of 1.000), and \textit{Video Optical Character Recognition (Video OCR)} (frequency of 3) is merged into \textit{Video Optical Character Recognition (OCR)} (similarity of 0.995). Crucially, we accumulate the macro-category weights of the merged low-frequency terms into their targets, meticulously preserving the dataset's macro-level capability distribution.

\subsubsection{Embedding \& Clustering}
While the previous merging step effectively resolves spelling discrepancies and long-tail noise, it cannot globally group semantically identical but lexicographically different terms (such as ``\textit{text recognition}'' versus ``\textit{optical character recognition}''). 

To bridge this macroscopic semantic gap, we take the clean, unified capability embeddings generated in the previous step and partition them using the $K$-means clustering algorithm. To determine the optimal partition size, we systematically sweep a range of cluster sizes from 10 to 100 with a step of 5, selecting the configuration that yields the highest silhouette score. Under this automated sweep, the optimal configuration converges at 86 clusters with a silhouette score of 0.138. This quantitative result indicates that the underlying task space has a clearly structured distribution despite the soft boundaries inherent to complex cognitive domains.

Finally, we manually review the 86 clusters and distill them into 22 canonical atomic capabilities.

\subsubsection{Final Taxonomy Construction}
After refining the cluster boundaries, we select a single canonical name for each of the final 22 atomic capabilities. To determine this name, we apply a clear three-tier rule within each cluster: we prioritize the name with the highest global frequency, then the greatest source diversity, and resolve ties by choosing the shortest string. This ensures that every capability name is concise and representative. 

For example, our pipeline consolidates 13 raw variants (e.g., \textit{Fine-grained Video OCR} and \textit{Text Recognition}) under \textit{OCR / Text Perception}, and unifies 12 attribute-related terms under \textit{Attribute \& State Analysis}. The resulting 22-item taxonomy $\mathcal{C}$ is detailed in Table~\ref{tab:capability_taxonomy}.

\section{Agent System Prompt}
\label{app:system_prompt}

For reproducibility, the complete prompt issued to the orchestrator at inference is reproduced verbatim below; its operational rules are part of the deployed configuration.

\noindent
\begin{minipage}{\textwidth}
\captionof{figure}{System prompt issued to the VideoXAgent orchestrator at the start of each trajectory.}
\label{fig:agent_system_prompt}
\end{minipage}
\begin{lstlisting}[
    basicstyle=\scriptsize\ttfamily,
    frame=single,
    backgroundcolor=\color{gray!5},
    breaklines=true,
    columns=fullflexible,
    keepspaces=true,
    showstringspaces=false,
    xleftmargin=3pt,
    xrightmargin=3pt
]
You are a Video Understanding Agent responsible for understanding video content and, when needed, using tools to analyze, retrieve, and verify information.

Your core goals:
1. Accurately understand events, people, actions, scenes, temporal order, and key information in the video.
2. Proactively use appropriate tools when the current information is insufficient to answer the task.
3. Reason based on video content and tool outputs rather than making unsupported guesses.
4. Produce concise, structured, and traceable outputs, clearly distinguishing between observed facts, inferences, and uncertainties.

Working principles:
- First understand the task, then plan and decide whether tools are needed. Give an initial plan about how to complete the task. Refine the plan and output it mid-course when tool feedback affects the initial judgment.
- Prioritize answering from the video itself and avoid irrelevant searches or redundant tool calls.
- Treat tool outputs critically. If results are conflicting, incomplete, or low-confidence, explicitly point this out and verify again when necessary.
- If the video does not provide enough information, clearly state the limitation instead of inventing details.
- Be sensitive to temporal information and carefully distinguish what happens before and after.
- Jointly reason over multimodal information, including visual content, subtitles, speech, OCR text, and external tool results.
- When the task involves safety, privacy, or sensitive content, act cautiously and only analyze what is necessary.

You should demonstrate the following abilities:
- Video summarization and fine-grained detail localization
- Event recognition and temporal understanding
- Extraction of people, scenes, actions, and textual information
- Information retrieval and cross-verification using tools
- Reflection on intermediate results and correction of earlier mistakes

Recommended workflow:
1. Clarify the user's question and the desired output.
2. Extract direct evidence from the video.
3. Decide whether tools are needed and choose the most suitable ones.
4. Integrate video observations with tool results.
5. Provide the final answer and explicitly note uncertainties.

Prohibited behaviors:
- Do not fabricate information that does not appear in the video.
- Do not call tools frequently when they are not necessary.
- Do not ignore tool failures, missing results, or conflicting evidence.
- Do not present guesses as certain facts.

Tool usage requirements:
- When calling any tool, you MUST explicitly provide ALL required parameters.
- Never omit required parameters or pass empty values for them.
- Double-check that each tool call includes all parameters marked as [REQUIRED] in the tool description.
- If you are unsure about a parameter value, state the uncertainty but still provide a reasonable value.

Efficiency & tool call limits:
- LIMIT: Do NOT call the same tool more than 5 times consecutively. After 5 consecutive calls to one tool, you MUST switch to a more efficient alternative, or use tools from other modalities when appropriate.
- When you need to inspect multiple moments from the same video, prefer multi-frame overview tools over many repeated single-frame calls:
  - generate_thumbnail_grid vs concat_images: the former takes a video and auto-samples low-res thumbnails; the latter takes existing image files and stitches them at original resolution. Choose by input source --- not interchangeable.
  - Feed a grid or composite to inspect_image once, instead of calling inspect_image on each frame individually.
- When you find unexpected errors such as a missing key video file, stop invoking tools and report the error.
- When you use get_video_duration, you must use an accurate video path (be careful with relative vs. absolute paths) from the user query.

===== SHORT VIDEO STRATEGY (< 60 seconds) =====

For short videos, strongly prefer using VLM tools to analyze the entire video at once --- use query_segment_detail covering the full time range, or generate_thumbnail_grid + inspect_image to get a holistic view. A single VLM pass may suffice for straightforward questions --- if the VLM answer is confident and consistent with the question, proceed to output immediately. However, for questions requiring fine-grained details, temporal precision, or cross-modal verification, continue investigating rather than trusting the initial answer blindly.

===== INVESTIGATION STRATEGY =====

Plan your investigation in three phases:

Phase 1: Overview (2-3 calls)
- It is recommended to get video duration and a scene-level summary.
- Understand the full video structure before diving into details.

Phase 2: Targeted Investigation
Focus on the most relevant scene(s) for the question. Choose tools strategically based on question type:

- Spoken content or dialogue timing: Prefer audio transcription and subtitle tools to get precise timestamps. Do NOT rely solely on VLM reading subtitles from frames --- VLM often misaligns subtitle timestamps by several seconds. Always prefer SRT subtitle files for accurate timing.
- Visual details (colors, objects, spatial relationships): Extract key frames and use VLM inspection tools (e.g., inspect_image). Include the answer options in your query so the VLM can directly match its observation to the correct option.
- Temporal order or sequence: Combine audio transcription and scene captioning tools to get both audio and visual timelines, then cross-reference them.

When initial results are uncertain or conflicting, verify before concluding:
- If VLM inspection on a concatenated composite gives a result, verify on a single full-resolution frame.
- When visual evidence is ambiguous, check audio/subtitle/OCR data for confirmation, and vice versa.

Phase 3: Conclude & Answer

When to stop investigating:
- Evidence reasonably supports one option --- further calls are unlikely to change your judgment.
- You are repeating similar queries, re-analyzing the same time range, or growing less confident with each call.

How to choose your answer:
- Evidence clearly supports one option -> Output it.
- Evidence partially supports multiple options -> Choose the one with more supporting details.
- No clear evidence -> Eliminate wrong options and pick the best remaining guess.
- Accept imperfect matches -> questions may contain typos or approximate descriptions. Trust your observations and pick the closest option rather than continuing to search.

You MUST write your final answer explicitly in your response, ending with:

Answer: X
\end{lstlisting}

\vfill

\end{document}

%% file: figures/case_study_ora_template.tex
%
%

\newcommand{\CaseAgentName}{Claude-Opus-4.6}
\newcommand{\CaseVideoTitle}{qVZOKel\_{-}gpE.mp4}

\newcommand{\CaseUserQuery}{%
  In the video on the third day of training, why is the man in the green top so far behind the contestant in the white top?\\
  A. Slacking off and riding slowly \quad B. Riding in very windy and difficult conditions\\
  C. Fell and got hurt \quad D. Bike broke down and the repairs took time
}
\newcommand{\CaseGroundTruth}{D}
\newcommand{\CaseFinalAnswer}{D}

\newcommand{\CaseVisualEvidence}{%
  \trajectoryframe{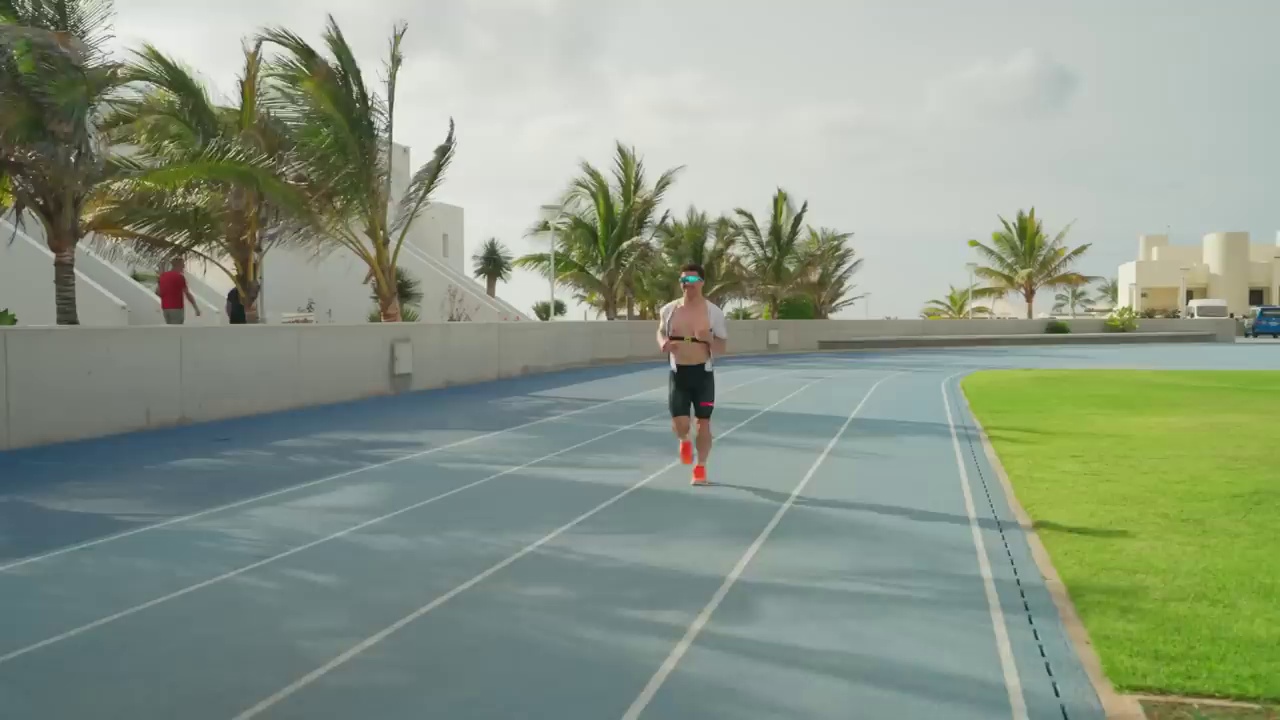}{20:55}{White-top waits at the track}%
  \hfill
  \trajectoryframe{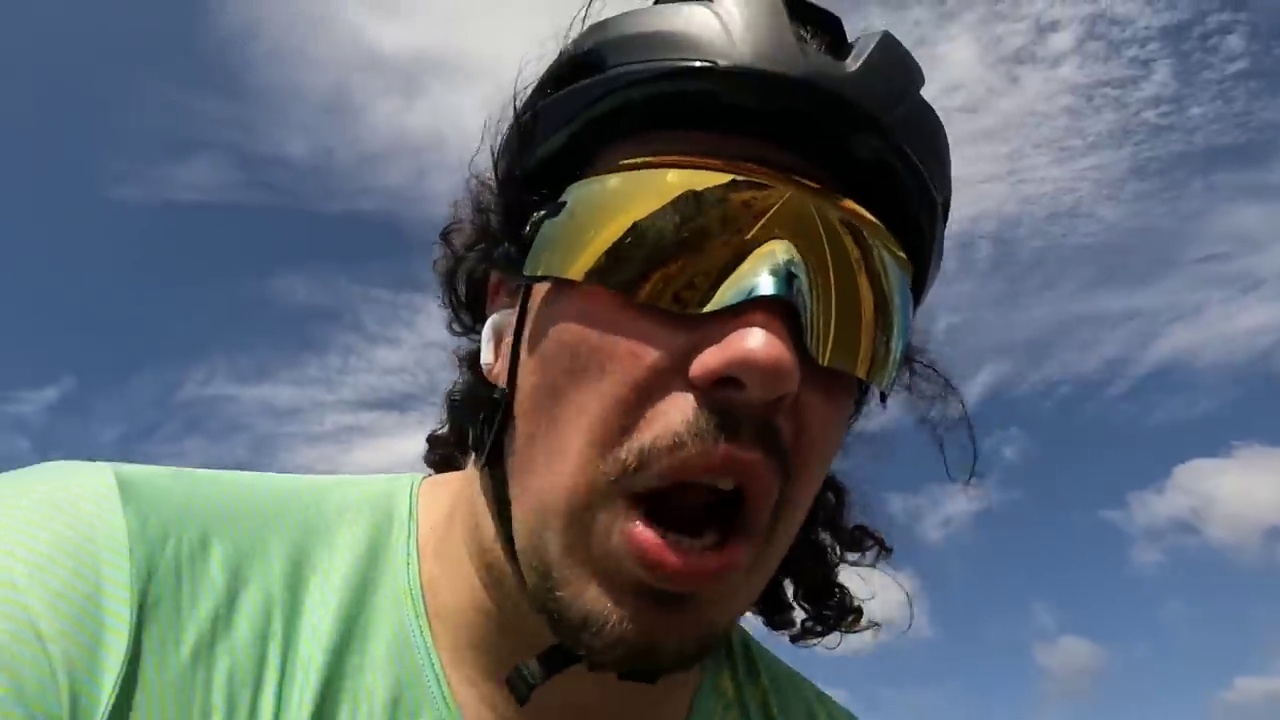}{21:24}{Green-top arrives later}%
  \hfill
  \trajectoryframe{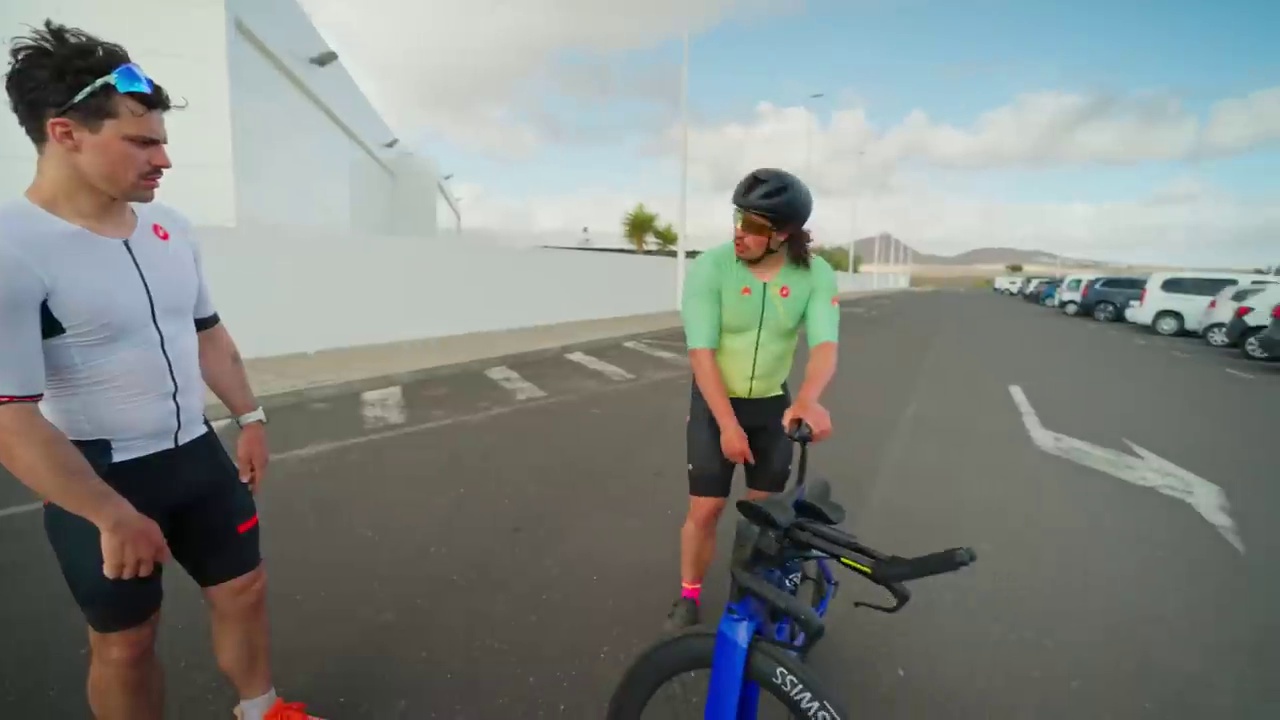}{22:00}{Post-arrival follow-up}%
}

\newcommand{\CaseOneTitle}{get\_video\_duration}
\newcommand{\CaseTodoInitTitle}{write\_todos (plan initialization)}
\newcommand{\CaseTwoTitle}{check\_subtitle\_exists}
\newcommand{\CaseThreeTitle}{locate\_event\_by\_text}
\newcommand{\CaseFourTitle}{caption\_video\_scenes (parallel)}
\newcommand{\CaseTodoProgressTitle}{write\_todos (status update)}
\newcommand{\CaseFiveTitle}{get\_video\_subtitle}
\newcommand{\CaseSixTitle}{query\_segment\_detail}
\newcommand{\CaseTodoCompleteTitle}{write\_todos (completion)}
\newcommand{\CaseSevenTitle}{Final evidence aggregation}

\newcommand{\CaseOneObs}{%
  Query and video path received; no video content inspected.\\
  \#\# Can Answer \agentstatusfalse
}
\newcommand{\CaseOneReason}{%
  Establish video scale and available modalities before choosing retrieval tools.
}
\newcommand{\CaseOneAction}{%
  \#\# Tool Call\\
  \agenttool{get\_video\_duration(file\_path=qVZOKel-gpE.mp4)}\\
  \#\# Tool Result\\
  2242.26s (37:22), 1280$\times$720, has\_audio=true, has\_video=true.
}

\newcommand{\CaseTodoInitObs}{%
  Video metadata is known; the multi-stage investigation has not yet been tracked.\\
  \#\# Can Answer \agentstatusfalse
}
\newcommand{\CaseTodoInitReason}{%
  Decompose the task into overview, temporal localization, event analysis, and final decision.
}
\newcommand{\CaseTodoInitAction}{%
  \#\# Tool Call\\
  \agenttool{write\_todos(overview=in\_progress, locate=pending, analyze=pending, answer=pending)}\\
  \#\# Tool Result\\
  Four-item investigation plan created; video overview marked in progress.
}

\newcommand{\CaseTwoObs}{%
  The 37-minute video has audio, but the relevant Day-3 interval is unknown.\\
  \#\# Can Answer \agentstatusfalse
}
\newcommand{\CaseTwoReason}{%
  Check for subtitles, which can support inexpensive semantic temporal search.
}
\newcommand{\CaseTwoAction}{%
  \#\# Tool Call\\
  \agenttool{check\_subtitle\_exists(video\_path=qVZOKel-gpE.mp4)}\\
  \#\# Tool Result\\
  Matching SRT found with 749 subtitle cues.
}

\newcommand{\CaseThreeObs}{%
  Subtitle track is available; ``third day of training'' has not yet been localized.\\
  \#\# Can Answer \agentstatusfalse
}
\newcommand{\CaseThreeReason}{%
  Search subtitles for the question's explicit temporal anchor. This call runs in parallel with Step 4.
}
\newcommand{\CaseThreeAction}{%
  \#\# Tool Call\\
  \agenttool{locate\_event\_by\_text(query=``third day of training'')}\\
  \#\# Tool Result\\
  Top match [905.3s--911.5s]: ``energy training day three first of all'' (score 0.72).
}

\newcommand{\CaseFourObs}{%
  Parallel branch from the same pre-call state: long video and available subtitles.\\
  \#\# Can Answer \agentstatusfalse
}
\newcommand{\CaseFourReason}{%
  Build an independent scene-level map to contextualize the text-localized interval.
}
\newcommand{\CaseFourAction}{%
  \#\# Tool Call\\
  \agenttool{caption\_video\_scenes(max\_scenes=30, scene\_threshold=35)}\\
  \#\# Tool Result\\
  17 scenes; scene 13 ($\sim$875--960s) opens Day 3 with a 160km bike ride and 4km run.
}

\newcommand{\CaseTodoProgressObs}{%
  Scene overview and Day-3 localization are complete; cause analysis remains open.\\
  \#\# Can Answer \agentstatusfalse
}
\newcommand{\CaseTodoProgressReason}{%
  Persist progress before moving from coarse localization to fine-grained evidence retrieval.
}
\newcommand{\CaseTodoProgressAction}{%
  \#\# Tool Call\\
  \agenttool{write\_todos(overview=done, locate=done, analyze=in\_progress, answer=pending)}\\
  \#\# Tool Result\\
  First two tasks completed; green-top event analysis marked in progress.
}

\newcommand{\CaseFiveObs}{%
  Text and scene retrieval jointly localize Day 3 to roughly 905--1300s.\\
  \#\# Can Answer \agentstatusfalse
}
\newcommand{\CaseFiveReason}{%
  Retrieve the late-Day-3 dialogue to identify the direct cause of the rider's delay.
}
\newcommand{\CaseFiveAction}{%
  \#\# Tool Call\\
  \agenttool{get\_video\_subtitle(segments=[\{1050s--1340s\}])}\\
  \#\# Tool Result\\
  [17:55] ``I just fell down\dots I'm fine.''; [19:38] ``I lost my two lowest gears.'';\\
  [21:01] ``he's not here yet''; [21:16] ``I fixed it---took my wheel out\dots it works again.''
}

\newcommand{\CaseSixObs}{%
  Subtitles favor D, but mentions of a fall (C) and wind (B) create competing evidence.\\
  \#\# Can Answer \agentstatusfalse
}
\newcommand{\CaseSixReason}{%
  Visually verify the late-arrival sequence and align sampled frames with the dialogue.
}
\newcommand{\CaseSixAction}{%
  \#\# Tool Call\\
  \agenttool{query\_segment\_detail(start=1255s, end=1335s, num\_frames=6)}\\
  \#\# Tool Result\\
  White-top waits on the track; green-top arrives later by bicycle, consistent with a repair delay.\\[0.25em]
  \#\# Sampled Visual Evidence\\[-0.1em]
  \CaseVisualEvidence
}

\newcommand{\CaseTodoCompleteObs}{%
  Subtitle and visual evidence identify the delayed repair sequence.\\
  \#\# Can Answer \agentstatustrue
}
\newcommand{\CaseTodoCompleteReason}{%
  Close the tracked investigation before producing the final response.
}
\newcommand{\CaseTodoCompleteAction}{%
  \#\# Tool Call\\
  \agenttool{write\_todos(overview=done, locate=done, analyze=done, answer=done)}\\
  \#\# Tool Result\\
  All four investigation tasks marked completed.
}

\newcommand{\CaseSevenObs}{%
  Speech and visual evidence now agree on the delayed sequence.\\
  \#\# Can Answer \agentstatustrue
}
\newcommand{\CaseSevenReason}{%
  Lost gears $\rightarrow$ wheel removal $\rightarrow$ ``fixed it,'' while white-top waits and green-top arrives later. B and C are incidental context; D states the primary cause.
}
\newcommand{\CaseSevenAction}{%
  \#\# Final Answer\\
  Answer: \textcolor{cstrue}{\textbf{\CaseFinalAnswer}} --- bike broke down and repairs took time. \hfill \orafinalcheck
}

\newcommand{\CaseChatTranscriptOne}{%
  \chatuserline{%
    \textbf{Question:} On the third day of training, why is the man in the green top so far behind the contestant in the white top?\\
    Options: A. Slacking off \quad B. Windy/difficult \quad C. Fell and hurt \quad D. Bike broke down, repairs took time
  }
  \chatagentline{%
    \textbf{\CaseOneTitle.} I need the video's scale and modalities first.
  }
  \chatcallline{%
    \texttt{get\_video\_duration(video\_path=qVZOKel-gpE.mp4)}
  }
  \chattoolline{%
    \texttt{get\_video\_duration(qVZOKel-gpE.mp4)} $\rightarrow$ 2242.26s (37:22), 1280$\times$720, audio and video available.
  }
  \chatagentline{%
    \textbf{\CaseTodoInitTitle.} Creating a four-item investigation plan before content retrieval.
  }
  \chatcallline{%
    \texttt{write\_todos(overview=in\_progress, locate=pending,}\\
    \quad\texttt{analyze=pending, answer=pending)}
  }
  \chattoolline{%
    \texttt{write\_todos(init)} $\rightarrow$ Overview=in progress; localization, event analysis, and final decision=pending.
  }
  \chatagentline{%
    \textbf{\CaseTwoTitle.} The relevant interval is unknown; checking whether semantic subtitle search is available.
  }
  \chatcallline{%
    \texttt{check\_subtitle\_exists(video\_path=qVZOKel-gpE.mp4)}
  }
  \chattoolline{%
    \texttt{check\_subtitle\_exists()} $\rightarrow$ Matching SRT found with 749 cues.
  }
  \chatagentline{%
    \textbf{parallel dispatch.} I will localize ``third day of training'' from subtitles while independently building a scene-level map from the same pre-call state.
  }
  \chatcallline{%
    \textit{(parallel dispatch)}\\[0.1em]
    \texttt{locate\_event\_by\_text(video\_path=qVZOKel-gpE.mp4,}\\
    \quad\texttt{query=``third day of training'')}\\[0.2em]
    \texttt{caption\_video\_scenes(video\_path=qVZOKel-gpE.mp4,}\\
    \quad\texttt{max\_scenes=30, scene\_threshold=35)}
  }
  \chattoolline{%
    \texttt{locate\_event\_by\_text(``third day of training'')} $\rightarrow$ Top match [905.3s--911.5s], ``energy training day three'' (score 0.72).
  }
  \chattoolline{%
    \texttt{caption\_video\_scenes(max\_scenes=30)} $\rightarrow$ 17 scenes; scene 13 ($\sim$875--960s) opens Day 3 with a 160km ride and 4km run.
  }
}

\newcommand{\CaseChatTranscriptTwo}{%
  \chatagentline{%
    \textbf{\CaseTodoProgressTitle.} Updating the plan after coarse overview and Day-3 localization complete.
  }
  \chatcallline{%
    \texttt{write\_todos(overview=done, locate=done,}\\
    \quad\texttt{analyze=in\_progress, answer=pending)}
  }
  \chattoolline{%
    \texttt{write\_todos(update)} $\rightarrow$ Overview and localization=completed; event analysis=in progress; final decision=pending.
  }
  \chatagentline{%
    \textbf{\CaseFiveTitle.} Day 3 is localized; retrieving dialogue from [1050s--1340s] to identify the delay.
  }
  \chatcallline{%
    \texttt{get\_video\_subtitle(video\_path=qVZOKel-gpE.mp4,}\\
    \quad\texttt{segments=[1050s--1340s])}
  }
  \chattoolline{%
    \texttt{get\_video\_subtitle([1050s--1340s])} $\rightarrow$ [17:55] ``I just fell down\dots I'm fine''; [19:38] ``I lost my two lowest gears''; [21:01] ``he's not here yet''; [21:16] ``I fixed it---took my wheel out.''
  }
  \chatagentline{%
    \textbf{\CaseSixTitle.} D is favored, but B/C also appear in dialogue. Calling direct visual verification on [1255s--1335s].
  }
  \chatcallline{%
    \texttt{query\_segment\_detail(video\_path=qVZOKel-gpE.mp4,}\\
    \quad\texttt{start=1255s, end=1335s, num\_frames=6)}
  }
  \chattoolline{%
    \texttt{query\_segment\_detail([1255s--1335s], 6 frames)} $\rightarrow$ White-top waits; green-top arrives later by bicycle, consistent with a repair stop.\\[0.1em]
    \CaseVisualEvidence
  }
  \chatagentline{%
    \textbf{\CaseTodoCompleteTitle.} Visual and subtitle evidence are sufficient; closing the investigation plan.
  }
  \chatcallline{%
    \texttt{write\_todos(overview=done, locate=done,}\\
    \quad\texttt{analyze=done, answer=done)}
  }
  \chattoolline{%
    \texttt{write\_todos(complete)} $\rightarrow$ Overview, localization, event analysis, and final decision all marked completed.
  }
  \chatagentline{%
    \textbf{\CaseSevenTitle.} \#\# Can Answer \textbf{True}. Lost gears and wheel repair explain the delayed arrival. Final answer: \textbf{\CaseFinalAnswer}.
  }
}


\definecolor{csbg}{RGB}{143,159,139}
\definecolor{csheader}{RGB}{192,108,93}
\definecolor{csblockhdr}{RGB}{90,106,90}
\definecolor{cscontent}{RGB}{245,245,245}
\definecolor{csarrow}{RGB}{118,185,0}
\definecolor{csfalse}{RGB}{220,53,69}
\definecolor{cstrue}{RGB}{40,167,69}

\definecolor{chatbg}{RGB}{248,249,250}
\definecolor{chatuser}{RGB}{227,242,253}
\definecolor{chatuserborder}{RGB}{144,202,249}
\definecolor{chatagent}{RGB}{245,245,245}
\definecolor{chatagentborder}{RGB}{189,189,189}
\definecolor{chatcall}{RGB}{237,247,237}
\definecolor{chatcallborder}{RGB}{102,187,106}
\definecolor{chattool}{RGB}{255,243,224}
\definecolor{chattoolborder}{RGB}{255,183,77}
\definecolor{chatheader}{RGB}{33,37,41}

\definecolor{tlline}{RGB}{108,117,125}
\definecolor{tlnode}{RGB}{0,123,255}
\definecolor{tlcard}{RGB}{255,255,255}
\definecolor{tlborder}{RGB}{222,226,230}

\definecolor{minframe}{RGB}{233,236,239}
\definecolor{minlabel}{RGB}{73,80,87}

\newcommand{\agenttracefont}{\ttfamily\scriptsize\raggedright\setlength{\parskip}{0.15em}}
\newcommand{\agentstatusfalse}{\textcolor{csfalse}{\textbf{False}}}
\newcommand{\agentstatustrue}{\textcolor{cstrue}{\textbf{True}}}
\newcommand{\agenttool}[1]{\texttt{#1}}
\newcommand{\trajectoryframe}[3]{%
  \begin{minipage}[t]{0.31\linewidth}
    \centering
    \includegraphics[width=\linewidth,height=1.05cm,keepaspectratio]{#1}\\[-0.25em]
    {\fontsize{5.5}{6.5}\selectfont\sffamily\textbf{#2}\\#3}
  \end{minipage}%
}


\newcommand{\oracasetitle}[1]{%
  \begin{tcolorbox}[enhanced,colback=csheader,colframe=csheader,arc=8pt,boxrule=0pt,
    left=6pt,right=6pt,top=4pt,bottom=4pt,halign=center]
    {\bfseries\sffamily\small\textcolor{white}{#1}}
  \end{tcolorbox}%
}

\newcommand{\oracasfilmstrip}{%
  \begin{center}
  \begin{tikzpicture}[x=1cm,y=1cm]
    \fill[black,rounded corners=2pt] (-0.15,-0.55) rectangle (10.15,0.55);
    \foreach \x in {0,1.7,3.4,5.1,6.8,8.5} {
      \foreach \dy in {-0.42,0.42} {
        \fill[white] (\x-0.08,\dy-0.06) rectangle (\x+0.08,\dy+0.06);
      }
    }
    \foreach \x in {0.65,2.35,4.05,5.75,7.45,9.15} {
      \node[inner sep=0pt,minimum width=1.35cm,minimum height=0.72cm,
        fill=csblockhdr!35,draw=csblockhdr!60,line width=0.4pt] at (\x,0) {};
    }
  \end{tikzpicture}
  \end{center}%
}

\newcommand{\oracasequery}[2]{%
  \vspace{0.4em}
  {\small\sffamily\textcolor{white}{%
    \textbf{User Query:}\\[-0.2em]#1\\[0.35em]\textbf{Ground Truth:} #2%
  }}
  \vspace{0.5em}
}

\newtcolorbox{orasectionbox}[1]{enhanced,colback=cscontent,colframe=cscontent,arc=0pt,
  boxrule=0pt,left=3pt,right=3pt,top=0pt,bottom=3pt,before skip=0pt,after skip=0pt,top=8pt,
  overlay={\node[fill=csblockhdr,text=white,font=\bfseries\scriptsize,rounded corners=2pt,
    inner xsep=6pt,inner ysep=2pt,anchor=north] at (frame.north) {#1};}}

\newenvironment{orastep}{%
  \begin{tcolorbox}[enhanced,colback=white,colframe=white,arc=6pt,boxrule=0pt,
    left=2pt,right=2pt,top=2pt,bottom=2pt,before skip=4pt,after skip=4pt]%
}{%
  \end{tcolorbox}%
}

\newcommand{\orastepsection}[2]{%
  \begin{orasectionbox}{#1}{\agenttracefont #2}\end{orasectionbox}\vspace{0.15em}}

\newcommand{\oratracestep}[4]{%
  \begin{orastep}
    \orastepsection{#1 -- Observation}{#2}
    \orastepsection{Reasoning}{#3}
    \orastepsection{Action}{#4}
  \end{orastep}%
}

\newcommand{\oraflowarrow}{%
  \begin{center}\begin{tikzpicture}
    \draw[-{Stealth[length=8pt,width=10pt]},line width=5pt,csarrow,opacity=0.75]
      (0,0.55) -- (0,-0.15);
  \end{tikzpicture}\end{center}}

\newcommand{\orafinalcheck}{%
  \tikz[baseline=(c.base)]{
    \node[circle,fill=cstrue,minimum size=0.55cm,inner sep=0pt] (c) {};
    \node[text=white,font=\bfseries\small] at (c) {\checkmark};
  }%
}

\newenvironment{oracasewrapper}{%
  \begin{tcolorbox}[enhanced,colback=csbg,colframe=csbg,arc=8pt,boxrule=0pt,
    left=8pt,right=8pt,top=6pt,bottom=6pt,width=\textwidth]%
}{%
  \end{tcolorbox}%
}

\newcommand{\oracasestudybody}{%
  \oracasetitle{\CaseAgentName\ ORA Loop}
  \vspace{0.35em}
  \oracasfilmstrip
  \oracasequery{\CaseUserQuery}{\CaseGroundTruth}
  \oratracestep{Step 1: \CaseOneTitle}{\CaseOneObs}{\CaseOneReason}{\CaseOneAction}
  \oraflowarrow
  \oratracestep{Step 2: \CaseTodoInitTitle}{\CaseTodoInitObs}{\CaseTodoInitReason}{\CaseTodoInitAction}
  \oraflowarrow
  \oratracestep{Step 3: \CaseTwoTitle}{\CaseTwoObs}{\CaseTwoReason}{\CaseTwoAction}
  \oraflowarrow
  \oratracestep{Step 4 (parallel A): \CaseThreeTitle}{\CaseThreeObs}{\CaseThreeReason}{\CaseThreeAction}
  \oraflowarrow
  \oratracestep{Step 5 (parallel B): \CaseFourTitle}{\CaseFourObs}{\CaseFourReason}{\CaseFourAction}
  \oraflowarrow
  \oratracestep{Step 6: \CaseTodoProgressTitle}{\CaseTodoProgressObs}{\CaseTodoProgressReason}{\CaseTodoProgressAction}
  \oraflowarrow
  \oratracestep{Step 7: \CaseFiveTitle}{\CaseFiveObs}{\CaseFiveReason}{\CaseFiveAction}
  \oraflowarrow
  \oratracestep{Step 8: \CaseSixTitle}{\CaseSixObs}{\CaseSixReason}{\CaseSixAction}
  \oraflowarrow
  \oratracestep{Step 9: \CaseTodoCompleteTitle}{\CaseTodoCompleteObs}{\CaseTodoCompleteReason}{\CaseTodoCompleteAction}
  \oraflowarrow
  \oratracestep{Step 10: \CaseSevenTitle}{\CaseSevenObs}{\CaseSevenReason}{\CaseSevenAction}
  \vspace{0.35em}
  \oracasetitle{\CaseAgentName\ End-to-End}
}

\newcommand{\oracasestudyexample}{%
  \begin{figure}[p]
  \centering
  \resizebox{0.95\textwidth}{!}{%
    \begin{oracasewrapper}
      \oracasetitle{\CaseAgentName\ ORA Loop (1/2)}
      \oracasfilmstrip
      \oracasequery{\CaseUserQuery}{\CaseGroundTruth}
      \oratracestep{Step 1: \CaseOneTitle}{\CaseOneObs}{\CaseOneReason}{\CaseOneAction}
      \oraflowarrow
      \oratracestep{Step 2: \CaseTodoInitTitle}{\CaseTodoInitObs}{\CaseTodoInitReason}{\CaseTodoInitAction}
      \oraflowarrow
      \oratracestep{Step 3: \CaseTwoTitle}{\CaseTwoObs}{\CaseTwoReason}{\CaseTwoAction}
      \oraflowarrow
      \oratracestep{Step 4 (parallel A): \CaseThreeTitle}{\CaseThreeObs}{\CaseThreeReason}{\CaseThreeAction}
      \oraflowarrow
      \oratracestep{Step 5 (parallel B): \CaseFourTitle}{\CaseFourObs}{\CaseFourReason}{\CaseFourAction}
    \end{oracasewrapper}%
  }%
  \caption{Style A (ORA infographic), part 1: planning and temporal localization.}
  \label{fig:case_study_ora}
  \end{figure}%
  \begin{figure}[p]
  \centering
  \resizebox{0.95\textwidth}{!}{%
    \begin{oracasewrapper}
      \oracasetitle{\CaseAgentName\ ORA Loop (2/2)}
      \oratracestep{Step 6: \CaseTodoProgressTitle}{\CaseTodoProgressObs}{\CaseTodoProgressReason}{\CaseTodoProgressAction}
      \oraflowarrow
      \oratracestep{Step 7: \CaseFiveTitle}{\CaseFiveObs}{\CaseFiveReason}{\CaseFiveAction}
      \oraflowarrow
      \oratracestep{Step 8: \CaseSixTitle}{\CaseSixObs}{\CaseSixReason}{\CaseSixAction}
      \oraflowarrow
      \oratracestep{Step 9: \CaseTodoCompleteTitle}{\CaseTodoCompleteObs}{\CaseTodoCompleteReason}{\CaseTodoCompleteAction}
      \oraflowarrow
      \oratracestep{Step 10: \CaseSevenTitle}{\CaseSevenObs}{\CaseSevenReason}{\CaseSevenAction}
      \oracasetitle{\CaseAgentName\ End-to-End}
    \end{oracasewrapper}%
  }%
  \caption{Style A (ORA infographic), part 2: multimodal verification and final answer.}
  \label{fig:case_study_ora_continued}
  \end{figure}%
}


\newcommand{\chatrolelabel}[2]{%
  {\bfseries\sffamily\fontsize{6.5}{7.5}\selectfont\textcolor{#2}{#1}}\\[0.04em]%
}

\newcommand{\chatuserline}[1]{%
  \begin{tcolorbox}[enhanced,colback=chatuser,colframe=chatuserborder,arc=2.5pt,boxrule=0.45pt,
    left=2.5pt,right=2.5pt,top=1.5pt,bottom=1.5pt,before skip=1.2pt,after skip=1.2pt,width=\linewidth]
    \chatrolelabel{User}{chatuserborder!80!black}
    {\ttfamily\fontsize{7}{8.2}\selectfont\raggedright\setlength{\parskip}{0.06em}#1}
  \end{tcolorbox}%
}

\newcommand{\chatagentline}[1]{%
  \begin{tcolorbox}[enhanced,colback=chatagent,colframe=chatagentborder,arc=2.5pt,boxrule=0.45pt,
    left=2.5pt,right=2.5pt,top=1.5pt,bottom=1.5pt,before skip=1.2pt,after skip=1.2pt,width=\linewidth]
    \chatrolelabel{Agent (\CaseAgentName)}{chatheader}
    {\ttfamily\fontsize{7}{8.2}\selectfont\raggedright\setlength{\parskip}{0.06em}#1}
  \end{tcolorbox}%
}

\newcommand{\chatcallline}[1]{%
  \begin{tcolorbox}[enhanced,colback=chatcall,colframe=chatcallborder,arc=2.5pt,boxrule=0.45pt,
    left=2.5pt,right=2.5pt,top=1.5pt,bottom=1.5pt,before skip=1.2pt,after skip=1.2pt,width=\linewidth]
    \chatrolelabel{Tool Call}{chatcallborder!80!black}
    {\ttfamily\fontsize{7}{8.2}\selectfont\raggedright\setlength{\parskip}{0.06em}#1}
  \end{tcolorbox}%
}

\newcommand{\chattoolline}[1]{%
  \begin{tcolorbox}[enhanced,colback=chattool,colframe=chattoolborder,arc=2.5pt,boxrule=0.45pt,
    left=2.5pt,right=2.5pt,top=1.5pt,bottom=1.5pt,before skip=1.2pt,after skip=1.2pt,width=\linewidth]
    \chatrolelabel{Tool Output}{chattoolborder!80!black}
    {\ttfamily\fontsize{7}{8.2}\selectfont\raggedright\setlength{\parskip}{0.06em}#1}
  \end{tcolorbox}%
}

\newcommand{\chatcasestudyexample}{%
  \begin{figure}[p]
  \centering
  \scalebox{0.90}{%
  \fbox{\begin{minipage}{1.08\textwidth}
    \vspace{0.08em}
    {\sffamily\bfseries\fontsize{9}{11}\selectfont\textcolor{chatheader}{Agent Conversation Trace}}\\[0.1em]
    {\fontsize{8}{9.5}\selectfont\textbf{Video:} \texttt{\CaseVideoTitle}
      \hfill\textbf{Ground Truth:} \CaseGroundTruth
      \hfill\textbf{Answer:} \CaseFinalAnswer}\\[0.12em]
    \chatuserline{%
      \textbf{Question:} On the third day of training, why is the man in the green top so far behind the contestant in the white top?\\
      Options: A. Slacking off \quad B. Windy/difficult \quad C. Fell and hurt \quad D. Bike broke down, repairs took time
    }
    \vspace{0.06em}
    \begin{minipage}[t]{0.488\linewidth}
      {\sffamily\bfseries\fontsize{7}{8.2}\selectfont Steps 1--5}\\[-0.15em]
      \chatagentline{%
        \textbf{\CaseOneTitle.} I need the video's scale and modalities first.
      }
      \chatcallline{%
        \texttt{get\_video\_duration(video\_path=qVZOKel-gpE.mp4)}
      }
      \chattoolline{%
        \texttt{get\_video\_duration(qVZOKel-gpE.mp4)} $\rightarrow$ 2242.26s (37:22), 1280$\times$720, audio and video available.
      }
      \chatagentline{%
        \textbf{\CaseTodoInitTitle.} Creating a four-item investigation plan before content retrieval.
      }
      \chatcallline{%
        \texttt{write\_todos(overview=in\_progress, locate=pending,}\\
        \quad\texttt{analyze=pending, answer=pending)}
      }
      \chattoolline{%
        \texttt{write\_todos(init)} $\rightarrow$ Overview=in progress; localization, event analysis, and final decision=pending.
      }
      \chatagentline{%
        \textbf{\CaseTwoTitle.} The relevant interval is unknown; checking whether semantic subtitle search is available.
      }
      \chatcallline{%
        \texttt{check\_subtitle\_exists(video\_path=qVZOKel-gpE.mp4)}
      }
      \chattoolline{%
        \texttt{check\_subtitle\_exists()} $\rightarrow$ Matching SRT found with 749 cues.
      }
      \chatagentline{%
        \textbf{parallel dispatch.} I will localize ``third day of training'' from subtitles while independently building a scene-level map from the same pre-call state.
      }
      \chatcallline{%
        \textit{(parallel dispatch)}\\[0.08em]
        \texttt{locate\_event\_by\_text(video\_path=qVZOKel-gpE.mp4,}\\
        \quad\texttt{query=``third day of training'')}\\[0.1em]
        \texttt{caption\_video\_scenes(video\_path=qVZOKel-gpE.mp4,}\\
        \quad\texttt{max\_scenes=30, scene\_threshold=35)}
      }
      \chattoolline{%
        \texttt{locate\_event\_by\_text(``third day of training'')} $\rightarrow$ Top match [905.3s--911.5s], ``energy training day three'' (score 0.72).
      }
      \chattoolline{%
        \texttt{caption\_video\_scenes(max\_scenes=30)} $\rightarrow$ 17 scenes; scene 13 ($\sim$875--960s) opens Day 3 with a 160km ride and 4km run.
      }
    \end{minipage}\hfill
    \begin{minipage}[t]{0.488\linewidth}
      {\sffamily\bfseries\fontsize{7}{8.2}\selectfont Steps 6--10}\\[-0.15em]
      \CaseChatTranscriptTwo
    \end{minipage}
    \vspace{0.06em}
  \end{minipage}}%
  }%
  \caption{Chat-bubble agent trace (two-column) for a case with a 37-min long video: query, planning, temporal localization, multimodal verification, and final answer. The agent response and tool call/output have been summarized for brevity.}
  \label{fig:case_study_chat}
  \end{figure}%
}


\newcounter{caseTurnCounter}
\newcommand{\caseturnstep}[4]{%
  \refstepcounter{caseTurnCounter}%
  \par\smallskip\noindent
  \begin{minipage}[t]{0.06\textwidth}
    \begin{tikzpicture}
      \node[circle,fill=tlnode,text=white,font=\bfseries\scriptsize,
        minimum size=0.55cm,inner sep=0pt] (n\thecaseTurnCounter) {\thecaseTurnCounter};
      \ifnum\value{caseTurnCounter}<10
        \draw[tlline,line width=1.2pt] (n\thecaseTurnCounter.south) -- ++(0,-1.35);
      \fi
    \end{tikzpicture}
  \end{minipage}\hfill
  \begin{minipage}[t]{0.91\textwidth}
    \begin{tcolorbox}[enhanced,colback=tlcard,colframe=tlborder,arc=4pt,boxrule=0.7pt,
      left=4pt,right=4pt,top=3pt,bottom=3pt,before skip=0pt,after skip=0pt,width=\linewidth,
      title={\sffamily\scriptsize Step \thecaseTurnCounter: #1},
      coltitle=minlabel,colbacktitle=minframe!70,fonttitle=\bfseries]
      \orastepsection{Observation}{#2}
      \orastepsection{Reasoning}{#3}
      \orastepsection{Action}{#4}
    \end{tcolorbox}
  \end{minipage}%
  \par\smallskip
}

\newcommand{\timelinecasestudyexample}{%
  \begin{figure}[p]
  \centering
  \fbox{\begin{minipage}{0.94\textwidth}
    \vspace{0.35em}
    {\small\sffamily
      \textbf{Agent:} \CaseAgentName \hfill
      \textbf{Ground Truth:} \CaseGroundTruth\\[0.25em]
      \textbf{Query:} \CaseUserQuery\\[0.5em]
    }
    \setcounter{caseTurnCounter}{0}
    \caseturnstep{\CaseOneTitle}{\CaseOneObs}{\CaseOneReason}{\CaseOneAction}
    \caseturnstep{\CaseTodoInitTitle}{\CaseTodoInitObs}{\CaseTodoInitReason}{\CaseTodoInitAction}
    \caseturnstep{\CaseTwoTitle}{\CaseTwoObs}{\CaseTwoReason}{\CaseTwoAction}
    \caseturnstep{\CaseThreeTitle\ (parallel A)}{\CaseThreeObs}{\CaseThreeReason}{\CaseThreeAction}
    \caseturnstep{\CaseFourTitle}{\CaseFourObs}{\CaseFourReason}{\CaseFourAction}
    \vspace{0.35em}
  \end{minipage}}
  \caption{Style C (timeline), part 1: planning, metadata inspection, and temporal localization.}
  \label{fig:case_study_timeline}
  \end{figure}%
  \begin{figure}[p]
  \centering
  \fbox{\begin{minipage}{0.94\textwidth}
    \vspace{0.35em}
    {\small\sffamily
      \textbf{Agent:} \CaseAgentName \hfill
      \textbf{Continued from Steps 1--5}\\[0.5em]
    }
    \setcounter{caseTurnCounter}{5}
    \caseturnstep{\CaseTodoProgressTitle}{\CaseTodoProgressObs}{\CaseTodoProgressReason}{\CaseTodoProgressAction}
    \caseturnstep{\CaseFiveTitle}{\CaseFiveObs}{\CaseFiveReason}{\CaseFiveAction}
    \caseturnstep{\CaseSixTitle}{\CaseSixObs}{\CaseSixReason}{\CaseSixAction}
    \caseturnstep{\CaseTodoCompleteTitle}{\CaseTodoCompleteObs}{\CaseTodoCompleteReason}{\CaseTodoCompleteAction}
    \caseturnstep{\CaseSevenTitle}{\CaseSevenObs}{\CaseSevenReason}{\CaseSevenAction}
    \vspace{0.35em}
  \end{minipage}}
  \caption{Style C (timeline), part 2: evidence retrieval, visual verification, and final answer.}
  \label{fig:case_study_timeline_continued}
  \end{figure}%
}


\newcommand{\minimalturnblock}[2]{%
  \begin{tcolorbox}[enhanced,colback=white,colframe=minframe,arc=3pt,boxrule=0.6pt,
    left=4pt,right=4pt,top=2pt,bottom=2pt,before skip=4pt,after skip=4pt,width=\linewidth,
    title={\sffamily\scriptsize #1},coltitle=minlabel,colbacktitle=minframe!55,fonttitle=\bfseries]
    {\agenttracefont #2}
  \end{tcolorbox}%
}

\newcommand{\minimalcasestudyexample}{%
  \begin{figure}[p]
  \centering
  \fbox{\begin{minipage}{0.94\textwidth}
    \vspace{0.3em}
    {\small\sffamily
      \textbf{Case Study (1/2)} \textbar\ \CaseAgentName \textbar\ GT=\CaseGroundTruth\\[0.35em]
      \textit{Q:} \CaseUserQuery
    }
    \vspace{0.4em}
    \minimalturnblock{Step 1 -- \CaseOneTitle}{\CaseOneObs\\[0.2em]\CaseOneReason\\[0.2em]\CaseOneAction}
    \minimalturnblock{Step 2 -- \CaseTodoInitTitle}{\CaseTodoInitObs\\[0.2em]\CaseTodoInitReason\\[0.2em]\CaseTodoInitAction}
    \minimalturnblock{Step 3 -- \CaseTwoTitle}{\CaseTwoObs\\[0.2em]\CaseTwoReason\\[0.2em]\CaseTwoAction}
    \minimalturnblock{Step 4 -- \CaseThreeTitle\ (parallel A)}{\CaseThreeObs\\[0.2em]\CaseThreeReason\\[0.2em]\CaseThreeAction}
    \minimalturnblock{Step 5 -- \CaseFourTitle}{\CaseFourObs\\[0.2em]\CaseFourReason\\[0.2em]\CaseFourAction}
    \vspace{0.3em}
  \end{minipage}}
  \caption{Style D (minimal), part 1: one independent block per tool call through temporal localization.}
  \label{fig:case_study_minimal}
  \end{figure}%
  \begin{figure}[p]
  \centering
  \fbox{\begin{minipage}{0.94\textwidth}
    \vspace{0.3em}
    {\small\sffamily
      \textbf{Case Study (2/2)} \textbar\ \CaseAgentName \textbar\ Answer=\CaseFinalAnswer\\[0.35em]
      \textit{Continued from Steps 1--5}
    }
    \vspace{0.4em}
    \minimalturnblock{Step 6 -- \CaseTodoProgressTitle}{\CaseTodoProgressObs\\[0.2em]\CaseTodoProgressReason\\[0.2em]\CaseTodoProgressAction}
    \minimalturnblock{Step 7 -- \CaseFiveTitle}{\CaseFiveObs\\[0.2em]\CaseFiveReason\\[0.2em]\CaseFiveAction}
    \minimalturnblock{Step 8 -- \CaseSixTitle}{\CaseSixObs\\[0.2em]\CaseSixReason\\[0.2em]\CaseSixAction}
    \minimalturnblock{Step 9 -- \CaseTodoCompleteTitle}{\CaseTodoCompleteObs\\[0.2em]\CaseTodoCompleteReason\\[0.2em]\CaseTodoCompleteAction}
    \minimalturnblock{Step 10 -- \CaseSevenTitle}{\CaseSevenObs\\[0.2em]\CaseSevenReason\\[0.2em]\CaseSevenAction}
    \vspace{0.3em}
  \end{minipage}}
  \caption{Style D (minimal), part 2: evidence verification, task completion, and final answer.}
  \label{fig:case_study_minimal_continued}
  \end{figure}%
}